\documentclass[onecolumn]{fairmeta}
\usepackage[utf8]{inputenc}
\usepackage[T1]{fontenc}
\usepackage{amsmath,amssymb,amsthm}
\usepackage{url,booktabs,graphicx,longtable}
\usepackage{algorithm,algpseudocode,float}
\usepackage{enumitem,multirow,nicefrac}
\usepackage{amsmath,amsfonts,bm}

\def\eqref#1{equation~\ref{#1}}
\def\1{\bm{1}}

\DeclareMathAlphabet{\mathsfit}{\encodingdefault}{\sfdefault}{m}{sl}
\SetMathAlphabet{\mathsfit}{bold}{\encodingdefault}{\sfdefault}{bx}{n}

\newcommand{\m}[1]{{TOPS}}

\floatstyle{ruled}
\restylefloat{algorithm}
\algrenewcommand\algorithmicrequire{\textbf{Require:}}
\algrenewcommand\algorithmicensure{\textbf{Output:}}

\title{TOPS: Reshaping Flow Source and Transport for Multi-Task Robot Manipulation}
\author[1]{Tuo An}
\author[1]{Chuhao Zhou}
\author[1]{Jiahua Dong}
\author[1]{Jingliang Li}
\author[1]{Jiaqi Bai}
\author[1,\dagger]{Jindou Jia}
\author[1,\dagger]{Jianfei Yang}
\affiliation[1]{MARS Lab, Nanyang Technological University}
\contribution[\dagger]{Joint corresponding authors}
\correspondence{\email{jindou.jia@ntu.edu.sg}, \email{jianfei.yang@ntu.edu.sg}}
\metadata[Project site]{\url{https://ntumars.github.io/TOPS/}}
\hypersetup{
  pdftitle={TOPS: Reshaping Flow Source and Transport for Multi-Task Robot Manipulation},
  pdfauthor={Tuo An, Chuhao Zhou, Jiahua Dong, Jingliang Li, Jiaqi Bai, Jindou Jia, Jianfei Yang}
}

\abstract{
Flow-based policies for language-conditioned multi-task manipulation commonly share a single \textit{Gaussian} source prior and a common linear transport path across tasks.
However, reliably executing multiple tasks with distinct motion requirements
remains challenging for a shared policy.
We introduce \textbf{\m{}} (\textbf{T}ask-\textbf{O}riented \textbf{P}riors and \textbf{S}chedules), which adapts the flow source prior and interpolation schedule to each task without modifying the network architecture.
\m{} first constructs multimodal task sources from reverse-mapped expert action sequences, allowing language instructions to select where generation begins. Guided by source-target covariance mismatch, it then slows the transition toward expert actions early in denoising and accelerates it later along directions requiring expansion. Under a \textit{Gaussian} endpoint model, our analysis shows that this schedule adjustment can lower an upper bound on velocity-prediction risk. Experiments on LIBERO, LIBERO-Plus, and CALVIN demonstrate improved performance across in-distribution, perturbation and cross-environment settings, alongside better data efficiency. Using one integration step, \m{} raises LIBERO-10 success from 53.44\% to 69.67\% compared with ten-step Standard Flow Matching (FM), while reducing inference time to approximately one-sixth of the baseline. \m{} also yields consistent gains when transferred to the large-scale
flow-based $\pi_{0.5}$ policy across all four evaluated settings. Real-world experiments further demonstrate improved task success, including an increase from 20\% with standard FM to 80\% with \m{} on the upper-drawer opening task.
}

\begin{document}
\maketitle

\section{Introduction}

Real-world manipulation exhibits substantial behavioral diversity.
Different tasks require distinct motion patterns, and even a given
goal can be achieved through different contact strategies, motion
trajectories, and temporal execution patterns.
Diffusion and flow-matching
models~\citep{actionflow,diffusion_policy,riemannian}
capture such multimodal action distributions directly and have
become widely used in imitation learning.
This expressive power has also carried the generative paradigm
from task-specific visuomotor control to language-conditioned
multi-task and generalist robot policies~\citep{dnact,pi0,pi05}.
In this setting, a shared policy must accommodate the motion
requirements of different tasks and execute each task reliably.

However, a shared policy can perform well on some tasks yet struggle
to execute others reliably~\citep{baku,libero,disc}.
A shared policy must capture multiple action
modes within a task and substantially different action distributions across
tasks~\citep{discrete_policy,flexible_policy}.
Existing approaches introduce or redesign policy components to
improve multi-task performance, using feature
modulation~\citep{baku}, visual selection~\citep{otter},
instruction-generated parameters~\citep{disc}, and expert
routing~\citep{smodp}.
In flow policies~\citep{flow_matching}, the network must recover distinct
task-specific action distributions from a shared \textit{Gaussian} source
under a common linear transport path.
Before changing the policy architecture as in prior work, we step back to the generative process itself and ask: \textbf{could adapting
the source prior and transport to each task improve multi-task
performance} without changing the policy
architecture?
To investigate, we reverse a trained multi-task flow policy and trace
expert actions back to their source samples~\citep{reversal_steer}.

\begin{figure}
    \centering
    \setlength{\abovecaptionskip}{0.5pt}
    \vspace{-10pt}
    \includegraphics[width=1\linewidth]{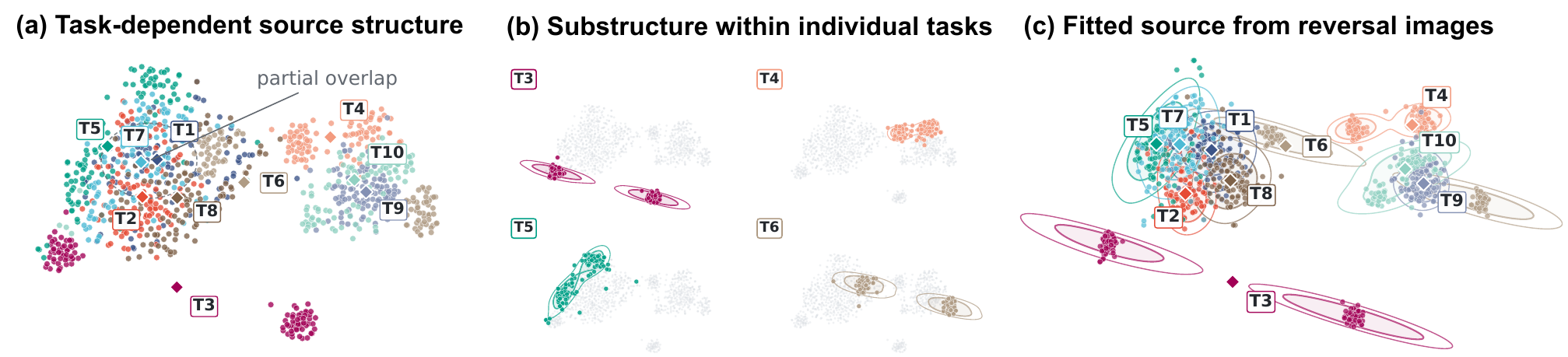}
    \vspace{-10pt}
    \caption{\textbf{Task-structured reverse images and the fitted source.} (a) t-SNE of reverse images from the ten LIBERO-10 tasks: tasks occupy distinct, partially overlapping regions. (b) The same embedding with one task highlighted per view, showing sub-clusters within individual tasks. (c) Samples from fitted per-task source in the same coordinates, reproducing the structure in (a).}
    \vspace{-5pt}
    \label{fig:raft_reverse_geometry}
\end{figure}

As shown in Figs.~\ref{fig:raft_reverse_geometry}(a,b), reverse-mapped expert actions exhibit task-dependent organization under the warm-up policy, with further subclusters within individual tasks. These patterns motivate us to investigate whether fitting task-specific sources to the reverse images and directly denoising from the corresponding region can improve policy learning.
To further understand what changes are needed to reach expert actions, we analyze the covariance mismatch between these sources and the corresponding expert-action distributions using a \textit{Gaussian} optimal-transport map. Its eigenvectors define orthogonal axes, each representing a weighted combination of action components across timesteps. We find expert actions have substantially greater standard deviation than source samples only along a few axes. We refer to this source-to-target increase as \textit{expansion}. Yet conventional flow matching~\citep{flow_matching,pi05} starts every task from a single task-independent \textit{Gaussian} and applies the same linear interpolation schedule along every axis. This leaves the network to recover task-specific action structure from shared noise, while interpolation timing remains unchanged despite the unequal source-to-target changes revealed by our analysis.

Motivated by these findings, we introduce \textbf{\m{}}
(\textbf{T}ask-\textbf{O}riented \textbf{P}riors and \textbf{S}chedules)
for multi-task flow policies.
It fits a \textit{Gaussian} mixture with low-rank covariance
structure~\citep{MPPCA} to each task's reverse-mapped samples,
allowing language instructions to select the source.
Along expansion axes, \m{} slows interpolation toward expert
actions early and accelerates it later, retaining linear
interpolation elsewhere.
Under a \textit{Gaussian} endpoint model~\citep{stochastic},
we show that adapting interpolation can lower an upper bound
on velocity-prediction risk.
This risk captures the irreducible error that arises when the
same interpolated action sequence corresponds to different
velocity targets.
On LIBERO-10, our schedule reduces estimated time-integrated
directional risk on expansion axes by $18.5\%$ relative to
linear interpolation (Fig.~\ref{fig:risk-validation}).
In controlled 2D experiments with few integration steps,
it also places more samples in target regions than linear
or fast-then-slow schedules (Fig.~\ref{fig:toy_exp}).
Together, these choices support efficient few-step action generation.

We evaluate \m{} on LIBERO~\citep{libero},
LIBERO-Plus~\citep{LIBERO-Plus}, CALVIN~\citep{calvin},
and three real-world multi-task suites.
With the same backbone as the baselines, \m{} achieves the highest
mean performance across two in-distribution and two out-of-distribution
settings, covering perturbations and unseen environments,
using one schedule-aware step.
Compared with ten-step Standard FM~\citep{flow_matching},
it raises LIBERO-10 success from 53.44\% to \textbf{69.67\%}
while reducing inference latency to approximately \textbf{one-sixth}.
Limited-data evaluations also show gains across all simulation
benchmarks: on CALVIN ABC$\to$D, \m{} trained on 30\% of
demonstrations outperforms all evaluated baselines trained on 50\%.
These gains also extend to the large-scale flow-based $\pi_{0.5}$ policy
in all four settings, raising LIBERO-Plus success from 64.4\%
to \textbf{69.0\%} and CALVIN ABC$\to$D average chain length
from 3.033 to \textbf{3.744}.
Real-world gains include \textbf{80\%} success on upper-drawer
opening, compared with 20\% for Standard FM.

To recap, our contributions are threefold:
\par\noindent\hspace{0.1cm}\hangindent=0.1cm\textbullet\ We construct
task-specific sources by fitting \textit{Gaussian} mixtures with
low-rank covariance structure to reverse-mapped expert action
chunks, enabling language-guided source selection.
\par\noindent\hspace{0.1cm}\hangindent=0.1cm\textbullet\ We introduce
slow-then-fast scheduling along expansion directions identified
by \textit{Gaussian} optimal transport, motivated by an upper
bound on velocity-prediction risk under independent
\textit{Gaussian} endpoints.
\par\noindent\hspace{0.1cm}\hangindent=0.1cm\textbullet\ Experiments
show that \m{} achieves the highest mean performance across the
main in- and out-of-distribution simulation settings with one
schedule-aware step, while reducing latency and improving data
efficiency. Gains also extend to the large-scale $\pi_{0.5}$
policy and real-world tasks.

\section{Related Work}

\paragraph{Language-conditioned multi-task policies.}
Language-conditioned multi-task policies aim to execute diverse
behaviors with a shared model~\citep{pi0,pi05,baku}.
Prior work supports task-specific behavior through visual feature
modulation or selection~\citep{baku,otter},
instruction-generated parameters~\citep{disc},
and semantic expert routing~\citep{smodp}.
\m{} adapts source distributions and interpolation schedules to
task-specific action structure while preserving the policy backbone
and its original conditioning.

\paragraph{Source distributions for generative policies.}
Existing sources use action histories~\citep{a2a,jia2026marspolicy,bai2026flash}
or conditioning representations~\citep{Cocos}, while LeaP~\citep{leap}
learns proprioception-conditioned diagonal \textit{Gaussians}.
MM-FM~\citep{MM-FM} combines a mixture fitted to target image latents
with mode-dependent coupling.
Flow reversal steering~\citep{reversal_steer} uses inverted actions
for direct steering or learning observation-conditioned noise policies
while keeping the base flow frozen.
\m{} fits low-rank \textit{Gaussian} mixtures to each task's reverse
images pooled across observations and trains the velocity
field under adapted sources and paths.

\paragraph{Coupling and path design in flow matching.}
Prior work optimizes endpoint couplings~\citep{OT-CFM,COT_policy}
or straightens trajectories by retraining on flow-generated
pairs~\citep{reflow}.
MeanFlow~\citep{meanflow} learns interval-averaged velocity for
one-step generation.
Other methods approximate geodesics under data-induced
metrics~\citep{metric_flow_maching} or adapt global schedules
to velocity profiles or regression difficulty~\citep{vsfm,dcfm}.
\m{} retains independent endpoint sampling and uses \textit{Gaussian}
OT to identify task-specific expansion directions for
slow-then-fast scheduling.

\section{Preliminaries and Motivating Analysis}
\label{sec:preliminaries}

\subsection{Flow-Matching Policies}
\label{sec:flow_matching_policies}

Let $\mathcal T$ denote the training tasks and
$\mathbf{c}=(o,\ell_\tau)$ the observation--instruction pair for
$\tau\in\mathcal T$; $\ell$ denotes an incoming instruction at inference.
An expert action chunk $\mathbf{x}_1\in\mathbb{R}^{D}$ contains
$H$ actions of dimension $d_a$, with $D=Hd_a$.
FM~\citep{flow_matching} trains a conditional velocity field to
predict $\mathbf{u}_t=\partial_t\mathbf{x}_t$ along a differentiable
path from $\mathbf{x}_0$ to $\mathbf{x}_1$:
\begin{equation}
    \mathcal{L}_{\mathrm{FM}}(\theta)
    =
    \mathbb{E}
    \left[
        \left\|
            \mathbf{v}_\theta(\mathbf{x}_t,t,\mathbf{c})-\mathbf{u}_t
        \right\|_2^2
    \right],
    \label{eq:fm_objective}
\end{equation}
where the expectation is over demonstration pairs
$(\mathbf{x}_1,\mathbf{c})$, with
$\mathbf{x}_1\sim p_{\mathrm{data}}(\cdot\mid\mathbf{c})$,
and independent draws $\mathbf{x}_0\sim p_0$ and
$t\sim\mathcal{U}(0,1)$.
Standard multi-task flow policies use a shared \textit{Gaussian}
source and linear interpolation:
\begin{equation}
    \mathbf{x}_0\sim\mathcal{N}(\mathbf{0},\sigma_0^2\mathbf{I}_D),
    \qquad
    \mathbf{x}_t=(1-t)\mathbf{x}_0+t\mathbf{x}_1,
    \qquad
    \mathbf{u}_t=\mathbf{x}_1-\mathbf{x}_0,
    \label{eq:standard_fm}
\end{equation}
where $\sigma_0>0$ is the source standard deviation and
$\mathbf{I}_D$ the $D$-dimensional identity matrix.
Inference integrates
$\dot{\mathbf{x}}(t)=\mathbf{v}_\theta(\mathbf{x}(t),t,\mathbf{c})$
from $\mathbf{x}(0)=\mathbf{x}_0$ to $t=1$, defining the forward
flow map $\Phi_{\theta,0\rightarrow1}^{\mathbf{c}}$.
Under standard regularity conditions, its inverse is
$\Phi_{\theta,1\rightarrow0}^{\mathbf{c}}
=(\Phi_{\theta,0\rightarrow1}^{\mathbf{c}})^{-1}$.

\subsection{Source Structure Revealed by Flow Reversal}
\label{sec:motivating_analysis}

With a shared \textit{Gaussian} source, where do different tasks'
expert actions map under flow reversal?
We examine a base policy $\mathbf{v}_{\theta_0}$ after a warm-up
with Eq.~(\ref{eq:standard_fm}), reversing each expert chunk under
its original observation and language condition $\mathbf{c}$:
\begin{equation}
    \widehat{\mathbf{x}}_0
    =
    \Phi_{\theta_0,1\rightarrow0}^{\mathbf{c}}(\mathbf{x}_1)
    \in\mathbb{R}^{D}.
    \label{eq:flow_reversal}
\end{equation}

For each task $\tau$, we pool reverse images across observations
into a bank $\mathcal{B}_\tau$.
The LIBERO-10 t-SNE~\citep{tsne} projection shows partially
overlapping task groups and within-task subclusters
(Figs.~\ref{fig:raft_reverse_geometry}(a,b)).
An intuitive explanation comes from the diversity of expert
behavior: action chunks differ across tasks and across execution
stages within a task, such as reaching, grasping, and placing.
The warm-up policy's reverse map may retain traces of these
behavioral differences, giving rise to the observed organization.
These observations motivate fitting a multimodal task-specific
source to each bank for training.
The source-only ablations assess the practical value of this
construction.

\begin{figure}
    \centering
    \setlength{\abovecaptionskip}{0.5pt}
    \includegraphics[width=0.9\linewidth]{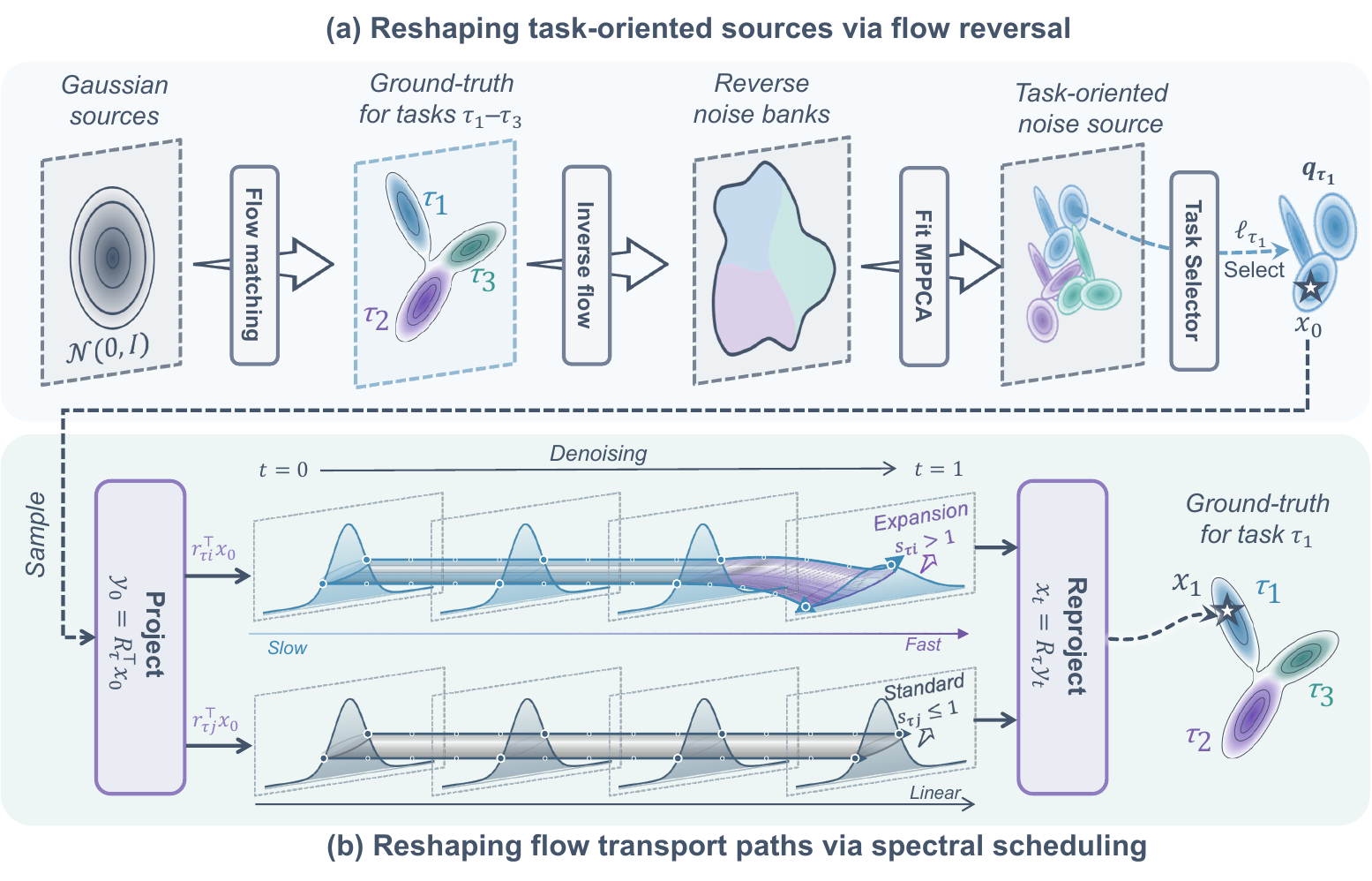}
    \caption{\textbf{Overview of \m{}.}
(a) We construct task-specific sources by fitting MPPCA models to reverse-mapped expert actions. An instruction selector maps the input language to an index \(\hat\tau\), identifying the source \(q_{\hat\tau}\) and its associated schedule. (b) In the selected task’s \textit{Gaussian} OT eigenbasis, we apply slow-then-fast interpolation along expansion directions and linear interpolation elsewhere, then map the interpolated states and velocity targets back to action space.
}
    \label{fig:main method}
\end{figure}

\section{Method}
\label{sec:method}
\subsection{Overview}
\label{sec:overview}

Building on the preceding analysis, \m{} adapts source distributions
and interpolation schedules to each task (Fig.~\ref{fig:main method}).
We fit multimodal sources to reverse-mapped expert actions and use
source--target covariance mismatch to guide slow-then-fast scheduling
along expansion axes, retaining linear interpolation elsewhere.
Both components are constructed offline and fixed during continuation
training of the velocity field.
At inference, a learned language selector chooses the source and
associated schedule, and a schedule-aware solver generates the action chunk.
The following subsections detail these components and their use in
training and inference.

\subsection{Task-Specific Source Construction}

The within-task subclusters in
Figs.~\ref{fig:raft_reverse_geometry}(a,b) motivate a multimodal
source for each task.
To capture this structure and correlations within action chunks,
we fit an $M_\tau$-component mixture of probabilistic principal
component analyzers (MPPCA)~\citep{MPPCA} to each task's
reverse-image bank $\mathcal{B}_\tau$:
\begin{equation}
\begin{aligned}
    q_\tau(\mathbf{x}_0)
    &=
    \sum_{m=1}^{M_\tau}
    \pi_{\tau m}\,
    \mathcal{N}
    \left(
        \mathbf{x}_0;
        \boldsymbol{\mu}_{\tau m},
        \boldsymbol{\Sigma}_{\tau m}
    \right),\\
    \boldsymbol{\Sigma}_{\tau m}
    &=
    \mathbf{U}_{\tau m}
    \boldsymbol{\Lambda}_{\tau m}
    \mathbf{U}_{\tau m}^{\top}
    + \sigma_{\tau m}^{2}\mathbf{I}_D.
\end{aligned}
\label{eq:task_source}
\end{equation}
Each component has weight $\pi_{\tau m}\geq0$, mean
$\boldsymbol{\mu}_{\tau m}\in\mathbb{R}^{D}$, and covariance
$\boldsymbol{\Sigma}_{\tau m}\in\mathbb{R}^{D\times D}$,
with $\sum_m\pi_{\tau m}=1$.
The orthonormal columns of
$\mathbf{U}_{\tau m}\in\mathbb{R}^{D\times r}$ define $r\ll D$
directions of coordinated variation across timesteps and action
dimensions.
The nonnegative diagonal matrix
$\boldsymbol{\Lambda}_{\tau m}\in\mathbb{R}^{r\times r}$
controls their strengths, while the isotropic residual with scale
$\sigma_{\tau m}>0$ allows variation beyond this subspace. We estimate the parameters offline using
expectation--maximization~\citep{EM}, with samples from the fitted
sources shown in Fig.~\ref{fig:raft_reverse_geometry}(c).
During training, each demonstration uses the source $q_\tau$
indexed by its task label $\tau$; at inference, the instruction
selector chooses the source
(Section~\ref{sec:raft_training_inference}).
The source provides a task-structured starting point, while the
conditional velocity field adapts generation to the current observation.

\subsection{Task-Specific Interpolation Scheduling}
\label{sec:raft_RAFT}

The task-specific source provides a structured starting point,
yet its variability and correlations may still differ from those
of expert actions.
We use this covariance mismatch to adapt interpolation timing
along different directions (Fig.~\ref{fig:main method}(b)).

\paragraph{Identifying expansion directions.}
For each task $\tau$, let
$\mathbf{C}_{\mathrm{s}}^\tau,
\mathbf{C}_{\mathrm{d}}^\tau\in\mathbb{R}^{D\times D}$
denote the source and expert covariances, computed analytically
from $q_\tau$ and estimated from demonstrations, respectively.
Both matrices are diagonally regularized to ensure positive
definiteness.
We relate them through the symmetric positive-definite
\textit{Gaussian} OT map
$\mathbf{T}_\tau\in\mathbb{R}^{D\times D}$:
\begin{equation}
\mathbf{T}_\tau
=
(\mathbf{C}_{\mathrm{s}}^\tau)^{-1/2}
\left[
(\mathbf{C}_{\mathrm{s}}^\tau)^{1/2}
\mathbf{C}_{\mathrm{d}}^\tau
(\mathbf{C}_{\mathrm{s}}^\tau)^{1/2}
\right]^{1/2}
(\mathbf{C}_{\mathrm{s}}^\tau)^{-1/2}.
\label{eq:raft_gaussian_ot}
\end{equation}
This map satisfies
$\mathbf{T}_\tau\mathbf{C}_{\mathrm{s}}^\tau\mathbf{T}_\tau
=\mathbf{C}_{\mathrm{d}}^\tau$.
Its eigendecomposition identifies directions along which
it acts by rescaling:
\begin{equation}
\mathbf{T}_\tau
=
\mathbf{R}_\tau
\operatorname{diag}(s_{\tau1},\ldots,s_{\tau D})
\mathbf{R}_\tau^\top.
\label{eq:tops_ot_eigenbasis}
\end{equation}
Each column $\mathbf{r}_{\tau j}$ of the orthogonal
$\mathbf{R}_\tau\in\mathbb{R}^{D\times D}$ represents a coordinated
action pattern across timesteps and action dimensions.
Covariance matching gives
$\mathbf{r}_{\tau j}^{\top}\mathbf{C}_{\mathrm{d}}^\tau
\mathbf{r}_{\tau j}
=s_{\tau j}^{2}
\mathbf{r}_{\tau j}^{\top}\mathbf{C}_{\mathrm{s}}^\tau
\mathbf{r}_{\tau j}$,
so $s_{\tau j}>0$ is the expert-to-source standard-deviation
ratio along this direction.
We therefore call directions with $s_{\tau j}>1$
\emph{expansion directions}.
Across our diagnostic settings, they occupy only $7.4$--$9.7\%$
of dimensions but account for $60.9$--$69.6\%$ of the
\textit{Gaussian}-OT covariance cost
(App.~\ref{app:expansion-analysis}).

\paragraph{Designing the interpolation schedule.}
For an expert action chunk $\mathbf{x}_1$ from task $\tau$,
we independently sample $\mathbf{x}_0\sim q_\tau$ and express
both endpoints in the OT eigenbasis:
$\mathbf{y}_i=\mathbf{R}_\tau^\top\mathbf{x}_i\in\mathbb{R}^{D}$,
$i\in\{0,1\}$.
Along direction $j$, we interpolate as
\begin{equation}
y_{t,j}
=
a_{\tau j}(t)y_{0,j}
+
[1-a_{\tau j}(t)]y_{1,j}.
\label{eq:raft_spectral_interpolation}
\end{equation}
Here, $a_{\tau j}:[0,1]\rightarrow[0,1]$ is a continuously
differentiable, nonincreasing source weight, with
$a_{\tau j}(0)=1$ and $a_{\tau j}(1)=0$.
As this weight decreases, the interpolant moves toward the
expert endpoint; linear interpolation uses $a_{\tau j}(t)=1-t$.

Near the source, interpolated states reveal little about expert
endpoints, leaving different velocity targets compatible with
similar states and conditioning.
Squared-loss regression averages these targets, and rapid
interpolation can amplify their differences along expansion
directions.
As the expert contribution grows, states become more informative,
motivating slower interpolation early and faster interpolation later.

To analyze this timing choice, we adopt a \textit{Gaussian}
surrogate within the stochastic-interpolant
framework~\citep{stochastic}.
The task-level source and expert marginals are modeled as
independent Gaussians with covariances
$\mathbf{C}_{\mathrm{s}}^\tau$ and $\mathbf{C}_{\mathrm{d}}^\tau$.
We study velocity prediction from the interpolated state alone,
defining its minimum achievable mean-squared error as
\emph{velocity-prediction risk}.
We upper-bound this risk by summing directional contributions.
Minimizing the integral of this bound over $t\in[0,1]$ yields
schedules $a_{\tau j}^{\star}$ satisfying
\begin{equation}
\frac{
|\dot{a}_{\tau j}^{\star}(1)|
}{
|\dot{a}_{\tau j}^{\star}(0)|
}
=
s_{\tau j},
\qquad j=1,\ldots,D.
\label{eq:raft_spectral_endpoint_ratio}
\end{equation}
Thus, along expansion directions, schedules minimizing this bound
transfer weight more slowly at the source endpoint than at the
expert endpoint.
App.~\ref{app:raft_risk_proof} provides the full bound and proof.

\paragraph{Constructing the path and velocity target.}
Guided by this endpoint asymmetry, we parameterize the source
weight with a bounded scheduling strength $\rho_{\tau j}$:
\begin{equation}
\begin{aligned}
\rho_{\tau j}
&=
\kappa\tanh(\gamma\log s_{\tau j})
\mathbf{1}_{\{s_{\tau j}>1\}},\\
a_{\tau j}(t)
&=
(1-t)e^{\rho_{\tau j}t},
\qquad \gamma>0,\quad 0<\kappa<1.
\end{aligned}
\label{eq:raft_retiming}
\end{equation}
Here, $\mathbf{1}_{\{\cdot\}}$ is the indicator function,
$\gamma$ controls sensitivity to $s_{\tau j}$, and $\kappa<1$
bounds the strength while keeping the source weight decreasing.
The resulting schedule is slow-then-fast along expansion
directions and linear elsewhere.

On LIBERO-10 expansion directions, this parameterization captures
$77.5\%$ of the improvement over linear interpolation attainable
by minimizing the time-integrated \textit{Gaussian} risk bound.
With fitted-source samples paired with recorded expert chunks,
it also reduces time-integrated velocity-prediction error on
this subspace by $22.1\%$ relative to linear interpolation,
as measured by matched MLPs using only the full interpolated
state and time
(Fig.~\ref{fig:risk-validation};
App.~\ref{app:risk-validation}).

We compute the basis and strengths offline and keep them fixed
during training.
Defining
\[
\mathbf{K}_\tau
=\mathbf{R}_\tau
\operatorname{diag}(-\rho_{\tau1},\ldots,-\rho_{\tau D})
\mathbf{R}_\tau^\top
\in\mathbb{R}^{D\times D},
\]
we obtain the interpolant and velocity target in action space:
\begin{equation}
\begin{aligned}
\mathbf{x}_t
&=
\mathbf{x}_1
+(1-t)e^{-t\mathbf{K}_\tau}
(\mathbf{x}_0-\mathbf{x}_1),\\
\mathbf{u}_t
=
\partial_t\mathbf{x}_t
&=
e^{-t\mathbf{K}_\tau}
\left[\mathbf{I}_D+(1-t)\mathbf{K}_\tau\right]
(\mathbf{x}_1-\mathbf{x}_0).
\end{aligned}
\label{eq:raft_RAFT}
\end{equation}
We train $\mathbf{v}_\theta$ to predict $\mathbf{u}_t$
using the FM loss.

\subsection{Training and Inference}
\label{sec:raft_training_inference}

We continue training from $\theta_0$ with $q_\tau$ and
$\mathbf{K}_\tau$ fixed.
We use the FM objective (Eq.~(\ref{eq:fm_objective})),
drawing $\mathbf{x}_0\sim q_\tau$ and constructing the interpolated
states and velocity targets according to Eq.~(\ref{eq:raft_RAFT}).

The lightweight language selector $\psi_\phi$ is trained offline
and remains fixed throughout policy training and evaluation.
At inference, it maps instruction $\ell$ to
$\hat{\tau}=\psi_\phi(\ell)$, selecting the corresponding
$q_{\hat{\tau}}$ and $\mathbf{K}_{\hat{\tau}}$.
We then sample $\mathbf{x}_0\sim q_{\hat{\tau}}$ and integrate
$\mathbf{v}_\theta$ with the associated schedule-aware solver.
By default, generation uses one two-stage step requiring two
velocity-network evaluations.
Appendices~\ref{app:open_language} and~\ref{app:integrator_details}
detail source selection and the solver, respectively;
Algorithm~\ref{alg:tops-pipeline} summarizes the full pipeline.

\newcommand{\sd}[1]{{\scriptsize$\pm$#1}}
\begin{table}[t]
\centering
\footnotesize
\setlength{\tabcolsep}{3.2pt}
\caption{\textbf{Main results and inference cost.}
LIBERO: success rate (\%); CALVIN: mean completed chain length
(out of five); higher is better for both.
$t$: per-call latency (ms).
Results are mean $\pm$ sample s.d. across three independent
training runs, after averaging evaluation repeats within each run.
LIBERO-Plus means cover 2,100 variants across six categories,
weighted by variant count; Camera is excluded because all
evaluated methods score zero.
Appendix~\ref{app:detailed} provides category and chain-depth breakdowns.}
\label{tab:main_results_compact}
\begin{tabular}{@{}lcccccc@{}}
\toprule
& \multicolumn{3}{c}{LIBERO} & \multicolumn{3}{c}{CALVIN} \\
\cmidrule(lr){2-4}\cmidrule(lr){5-7}
Method & ID & OOD & $t$ (ms) & D$\to$D & ABC$\to$D & $t$ (ms) \\
\midrule
Standard FM~\citep{flow_matching}
& 53.44\sd{2.93} & 27.11\sd{0.34} & 49.4
& 1.90\sd{0.05} & 0.70\sd{0.03} & 56.2 \\
\addlinespace[1pt]
OT-CFM~\citep{OT-CFM}
& 30.17\sd{2.72} & 19.02\sd{0.12} & 49.4
& 1.28\sd{0.05} & 0.46\sd{0.02} & 58.9 \\
2-ReFlow~\citep{reflow}
& 37.17\sd{3.12} & 20.22\sd{0.35} & 7.1
& 1.74\sd{0.05} & 0.66\sd{0.03} & 8.6 \\
RF-distill~\citep{reflow}
& 29.17\sd{0.94} & 17.56\sd{0.06} & 4.8
& 1.75\sd{0.05} & 0.70\sd{0.03} & 5.8 \\
\addlinespace[1pt]
MeanFlow~\citep{meanflow}
& 59.00\sd{2.83} & 34.73\sd{0.32} & 5.7
& 1.89\sd{0.02} & 0.86\sd{0.02} & 5.4 \\
PriorGrad~\citep{priorgrad}
& 62.33\sd{0.24} & 30.56\sd{0.30} & 48.3
& 1.97\sd{0.06} & \underline{1.02\sd{0.04}} & 55.4 \\
MM-FM~\citep{MM-FM}
& 34.67\sd{3.17} & 16.75\sd{0.49} & 48.8
& 1.75\sd{0.05} & 0.68\sd{0.03} & 52.5 \\
\addlinespace[1pt]
LeaP~\citep{leap}
& 60.25\sd{3.89} & 30.72\sd{0.83} & 49.6
& 1.74\sd{0.02} & 0.73\sd{0.03} & 58.7 \\
Regression~\citep{act}
& 40.17\sd{0.62} & 22.17\sd{0.50} & 5.1
& 0.88\sd{0.03} & 0.27\sd{0.02} & 5.7 \\
\midrule
Source only (fixed $M_\tau$)
& 59.17\sd{0.36} & \underline{35.21\sd{0.66}} & 7.6
& 2.12\sd{0.06} & 0.88\sd{0.04} & 8.8 \\
Source only (task-wise $M_\tau$)
& \underline{63.17\sd{0.64}} & 30.52\sd{0.66} & 7.6
& \underline{2.15\sd{0.06}} & 0.98\sd{0.04} & 8.6 \\
\textbf{\m{}}
& \textbf{69.67\sd{0.61}} & \textbf{35.29\sd{0.08}} & 8.2
& \textbf{2.27\sd{0.06}} & \textbf{1.05\sd{0.04}} & 9.0 \\
\bottomrule
\end{tabular}
\vspace{-5pt}
\end{table}

\begin{figure}
    \centering
    \setlength{\abovecaptionskip}{0.5pt}
    \includegraphics[width=1.0\linewidth]{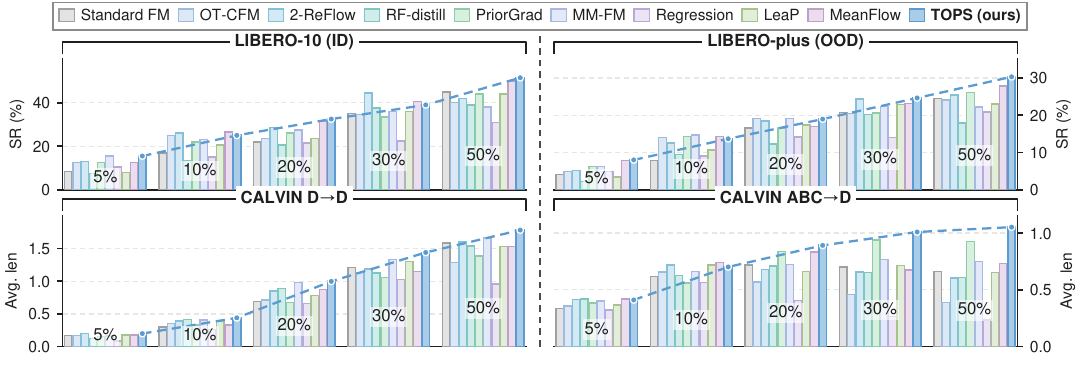}
    \vspace{-10pt}
    \caption{\textbf{Data efficiency.}
Performance with 5\%--50\% of the training demonstrations
on LIBERO-10, LIBERO-Plus, CALVIN D$\to$D, and CALVIN ABC$\to$D.
\m{} matches or exceeds the strongest baseline in 14 of
the 20 benchmark--data-budget configurations.}
    \vspace{-10pt}
    \label{fig:low_data results}
\end{figure}

\section{Experiments}
\label{sec:experiments}

\subsection{Simulation Experiments}
\label{sec:simulation_benchmarks}

\paragraph{Experimental setup and protocol.}
We evaluate in-distribution (ID) multi-task performance on
LIBERO-10~\citep{libero} and CALVIN D$\to$D~\citep{calvin},
and out-of-distribution (OOD) performance on
LIBERO-Plus~\citep{LIBERO-Plus} and CALVIN ABC$\to$D.
For ABC$\to$D, policies trained in environments A--C are tested
directly in unseen D.
All methods share the policy backbone, action representation,
and prediction horizon.
At inference, \m{} and its source-only variants use a learned
MiniLM--MLP~\citep{minilm} selector to choose sources from
input instructions.
On LIBERO-Plus, \m{} uses the provided instruction for both
source--schedule selection and policy conditioning, including
perturbed wording in the language category.
For Tab.~\ref{tab:main_results_compact}, \m{} uses one two-stage
schedule-aware step, while source-only variants and
2-ReFlow~\citep{reflow} use one explicit-midpoint step
(all NFE $=2$).
RF-distill~\citep{reflow} uses one explicit-Euler step and
MeanFlow~\citep{meanflow} its native one-step average-velocity
update (both NFE $=1$).
Standard FM~\citep{flow_matching}, OT-CFM~\citep{OT-CFM},
PriorGrad~\citep{priorgrad}, MM-FM~\citep{MM-FM}, and
LeaP~\citep{leap} use ten explicit-midpoint steps (NFE $=20$).
Regression~\citep{act} directly predicts action chunks.
We report LIBERO success rates, CALVIN average completed chain
lengths over 1,000 five-task chains unless otherwise specified,
and measured inference latency per policy call.
Implementation details appear in
App.~\ref{app:simulation_implementation}.

\subsubsection{\m{} improves multi-task performance with fewer sampling steps}

\paragraph{Best performance with few-step inference.}
With one solver step, \m{} achieves the highest mean performance
in both ID settings (Tab.~\ref{tab:main_results_compact}).
Compared with ten-step Standard FM, it raises LIBERO-10 success
from \textbf{53.44\%} to \textbf{69.67\%} and CALVIN D$\to$D
average completed chain length from \textbf{1.90} to \textbf{2.27}.
Per-call latency falls from \textbf{49.4} to \textbf{8.2} ms
on LIBERO and from \textbf{56.2} to \textbf{9.0} ms on CALVIN,
approximately one-sixth of Standard FM's latency in both settings.

\paragraph{Robustness and cross-environment generalization.}
The same one-step configuration yields the highest mean performance
in both OOD settings.
LIBERO-Plus success reaches \textbf{35.29\%}, compared with
27.11\%, 30.72\%, and 34.73\% for Standard FM, LeaP, and MeanFlow,
respectively.
On CALVIN ABC$\to$D, average completed chain length reaches
\textbf{1.05}, versus 0.70, 0.73, and 0.86 for the same baselines.
Tabs.~\ref{tab:liberoplus_categories} and~\ref{tab:calvin_chain}
provide category and chain-length breakdowns
(App.~\ref{app:detailed}).

\subsubsection{\m{} learns more effectively from fewer demonstrations}

Since \m{} fits task-specific sources from reversed expert demonstrations, this raises a natural question: \textit{can this source construction retain benefits when demonstrations are limited?}
To investigate this question, we train each method on 5\%, 10\%, 20\%, 30\%, and 50\% of the
demonstrations and evaluate the four ID and OOD settings using
the same protocol as above.
Fig.~\ref{fig:low_data results} shows that \m{} outperforms
Standard FM in all 20 configurations and matches or exceeds
the strongest baseline in 14.
With 5\% demonstrations, \m{} achieves \textbf{15.50\%}
success on LIBERO-10 and \textbf{8.00\%} on LIBERO-Plus,
compared with 8.50\% and 4.10\% for Standard FM, respectively.
On CALVIN ABC$\to$D, \m{} achieves average completed chain
length of \textbf{1.009} using 30\% of the demonstrations,
exceeding all baselines trained on 50\% (PriorGrad performs best at 0.925).

\subsubsection{TOPS Transfers to a Large-Scale Flow-Based VLA}

To test whether these gains extend to a larger flow-based policy,
we apply \m{} to $\pi_{0.5}$~\citep{pi05}, using the official
pretrained model and the same LoRA configuration as the baseline.
Both variants use the same total budget of 14K training steps;
\m{} resumes from an intermediate baseline checkpoint, changing
only the source distribution, transport target, and sampler.
At inference, the baseline uses ten Euler steps, while \m{} uses
one two-stage schedule-aware step (NFE $=2$).

\begin{figure}
    \centering
    \setlength{\abovecaptionskip}{0.5pt}
    \includegraphics[width=0.95\linewidth]{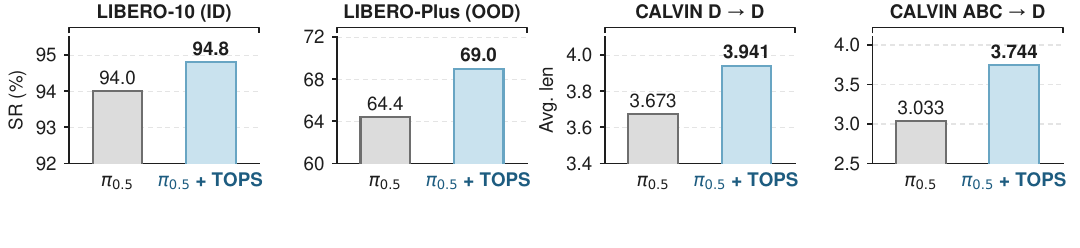}
    \vspace{-20pt}
    \caption{
\textbf{Transfer to the large-scale flow-based $\pi_{0.5}$ policy.}
Both variants use the same 14K-step training budget.
The baseline uses ten Euler inference steps, while \m{} uses one
schedule-aware step (NFE=2). Detailed results are provided in App.~\ref{app:pi05}.
}
    \vspace{-10pt}
    \label{fig:pi05_results}
\end{figure}

As shown in Fig.~\ref{fig:pi05_results}, \m{} improves $\pi_{0.5}$ in all
four evaluated settings. On CALVIN, the average completed chain length
increases from 3.673 to \textbf{3.941} on D$\rightarrow$D and from 3.033 to
\textbf{3.744} on ABC$\rightarrow$D. LIBERO-Plus success rises from 64.4\% to
\textbf{69.0\%}, while LIBERO-10 improves from an already strong 94.0\% to
\textbf{94.8\%}. These results show that reshaping the flow source and transport
remains beneficial when transferred to a substantially larger
flow-based policy, with particularly clear gains under distribution shift.
Detailed breakdowns are provided in App.~\ref{app:pi05}.

\begin{figure}
    \centering
    \setlength{\abovecaptionskip}{0.5pt}
    \includegraphics[width=1.0\linewidth]{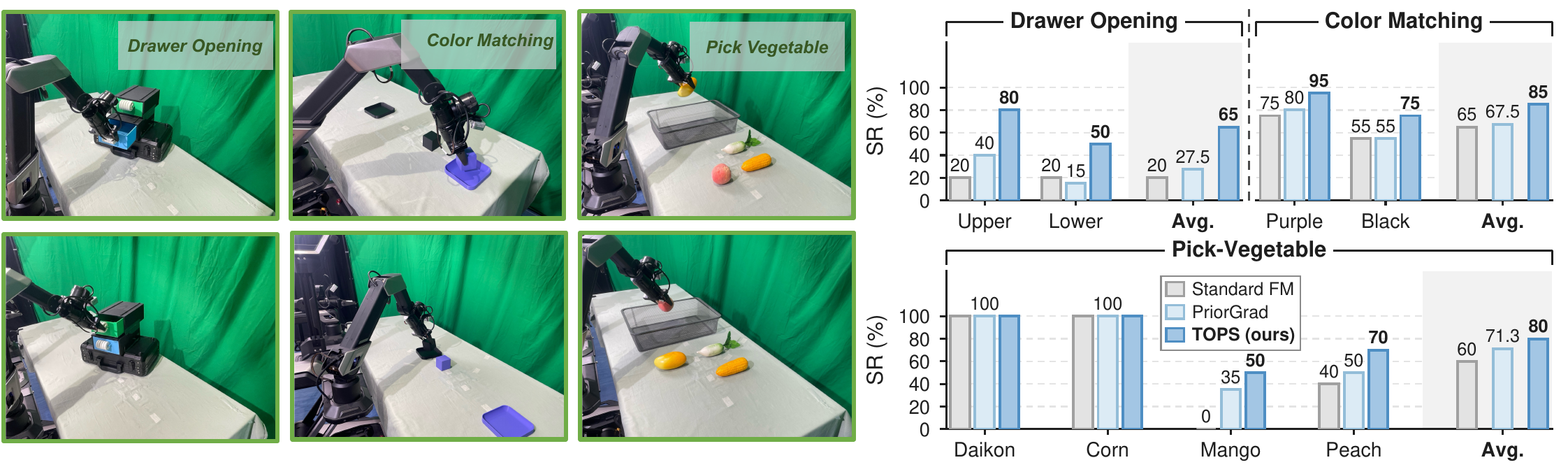}
    \vspace{-10pt}
    \caption{\textbf{Real-world evaluation.} Left: example scenes from the three task suites. Right: per-task success rates (\%) and suite averages (Avg.). Each task is evaluated over 20 rollouts.
}
    \vspace{-10pt}
    \label{fig:real_world_task_results}
\end{figure}

\subsection{Real-World Experiments}
\label{sec:real_world}

\paragraph{Experimental setup and protocol.}
We evaluate three multi-task suites on a Galaxea R1 Lite, actuating
only its right arm and two-finger gripper.
Policies receive RGB images from the head and wrist cameras,
proprioception, and language instructions.
Within each suite, all candidate targets are present in a shared
scene, and the instruction specifies the task.
Fig.~\ref{fig:real_world_task_results} shows the task setups.
\textbf{Drawer Opening} requires opening the instructed upper
or lower drawer.
\textbf{Color Matching} requires placing the instructed purple
or black cube into its matching tray.
\textbf{Pick-Vegetable} requires placing the instructed object
(daikon, corn, mango, or peach) in a collection box.
We collect 50 demonstrations per task for Drawer Opening and
Color Matching, and 30 for Pick-Vegetable.
We evaluate \m{}, Standard FM, and PriorGrad on eight tasks,
with 20 rollouts per task (160 per method).
We report per-task success rates and compare the pooled success
rates of \m{} and Standard FM using a two-sided two-proportion
$z$-test.
Implementation and statistical details appear in
App.~\ref{app:real-world-details}.

\paragraph{Multi-task manipulation performance.}
Fig.~\ref{fig:real_world_task_results} shows that \m{} matches both
baselines on daikon and corn (100\% success) and achieves higher
success on the other six tasks.
Compared with Standard FM, success on the upper and lower drawers
rises from 20\% each to \textbf{80\%} and \textbf{50\%}, respectively.
On Color Matching, \m{} reaches \textbf{95\%} and \textbf{75\%}
on the purple and black tasks, respectively, improving both by
20 percentage points over Standard FM.
On mango and peach, success increases from Standard FM's 0\% and
40\% to \textbf{50\%} and \textbf{70\%}, respectively.
Pooled across the eight tasks, \m{} succeeds in 124/160 rollouts
(77.50\%), compared with 82/160 (51.25\%) for Standard FM,
a statistically significant improvement of 26.25 percentage points
($z=4.90$, $p<0.001$).

\begin{figure}
    \centering
    \setlength{\abovecaptionskip}{0.5pt}
    \includegraphics[width=1.0\linewidth]{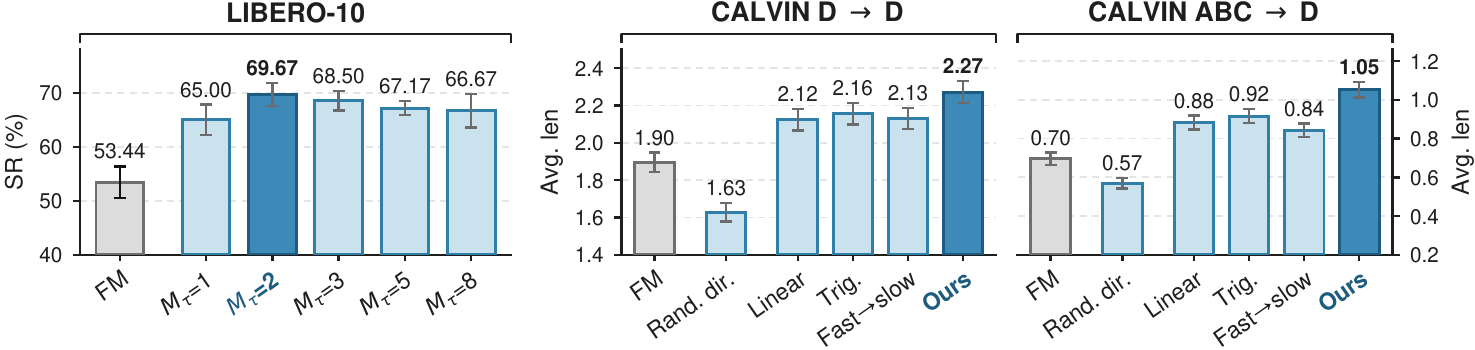}
    \vspace{-10pt}
    \caption{\textbf{Ablations of \m{}.}
Left: LIBERO-10 success (\%) versus component count $M_\tau$.
Middle and right: path ablations on CALVIN D$\to$D and
ABC$\to$D (mean completed chain length).
FM denotes Standard FM; Rand.\ dir.\ replaces expansion directions
with random ones; Trig.\ uses trigonometric interpolation.
The dark bar denotes full \m{}.
Error bars: LIBERO-10 s.d.\ across seeds (training for FM;
evaluation for $M_\tau$ variants) and CALVIN s.e.m.\ over
1,000 chains.}
    \vspace{-10pt}
    \label{fig:ablation_study}
\end{figure}

\subsection{Ablation Study}
\label{sec:ablation}

\paragraph{Do both components contribute?}
As shown in Tab.~\ref{tab:main_results_compact}, source-only variants with fixed $M_\tau=2$ or task-wise mixture
counts outperform Standard FM in all four simulation settings
under the same training budget and the reported inference budgets.
With the $M_\tau=2$ source unchanged, nonlinear scheduling and
its integrator jointly raise LIBERO-10 success from 59.17\% to 69.67\%.
Robomimic~\citep{robomimic} provides a complementary comparison
with a fixed \textit{Gaussian} source
(App.~\ref{app:robomimic_single_task}), while
App.~\ref{app:solver_ablation} compares solvers at a fixed checkpoint.

\paragraph{How sensitive is \m{} to the number of mixture components?}
We vary $M_\tau\in\{1,2,3,5,8\}$ in full \textsc{\m{}} on LIBERO-10,
using the same count across tasks and fixing covariance rank,
source scale, and training budget.
All tested values outperform Standard FM
(Fig.~\ref{fig:ablation_study}).
$M_\tau\in\{2,3\}$ yields the strongest results, with modest
degradation for the other tested counts.
This preference for small mixtures is consistent with the within-task
structure in Fig.~\ref{fig:raft_reverse_geometry} and supports our
default $M_\tau=2$ on LIBERO-10.

\paragraph{Do directions and scheduling matter?}
With the source fixed, we test random directions, shared
trigonometric interpolation, and fast-then-slow timing along
expansion directions.
Random directions substantially reduce performance in both
CALVIN settings (Fig.~\ref{fig:ablation_study}).
In a controlled 2D experiment, slow-then-fast timing also improves
few-step generation over linear and fast-then-slow schedules
(Fig.~\ref{fig:toy_exp}).
Matched LIBERO-10 MLPs without observation conditioning show
$22.1\%$ lower time-integrated velocity-prediction error on expansion
directions with slow-then-fast timing and $28.0\%$ higher error
with fast-then-slow timing, both relative to linear interpolation
(App.~\ref{app:risk-validation}).

\section{Conclusion}
\m{} incorporates task structure into source priors and interpolation
schedules while preserving the policy backbone. Reverse-mapped expert
actions guide multimodal source fitting, and an upper bound on
velocity-prediction risk under independent \textit{Gaussian} endpoints
motivates slow-then-fast scheduling along expansion directions.
Experiments demonstrate improved multi-task performance, robustness,
and data efficiency with one inference step, together with
higher real-world task success.
\paragraph{Limitations.}
\m{} incurs an offline cost to construct task-specific sources and
schedules, which remain fixed during subsequent training and
deployment. It selects among predefined tasks and uses a fixed
action-chunk length; generalization to unseen task identities is
outside our evaluation. Future work could study incremental source
and schedule updates and variable-horizon action generation.

\clearpage

\bibliography{iclr2027_conference}
\bibliographystyle{plainnat}

\clearpage
\appendix
\section{Appendix}
\subsection{Comprehensive evaluation results}
\label{app:detailed}

\begin{figure}[htbp]
    \centering
    \setlength{\abovecaptionskip}{0.5pt}
    \includegraphics[width=1\linewidth]{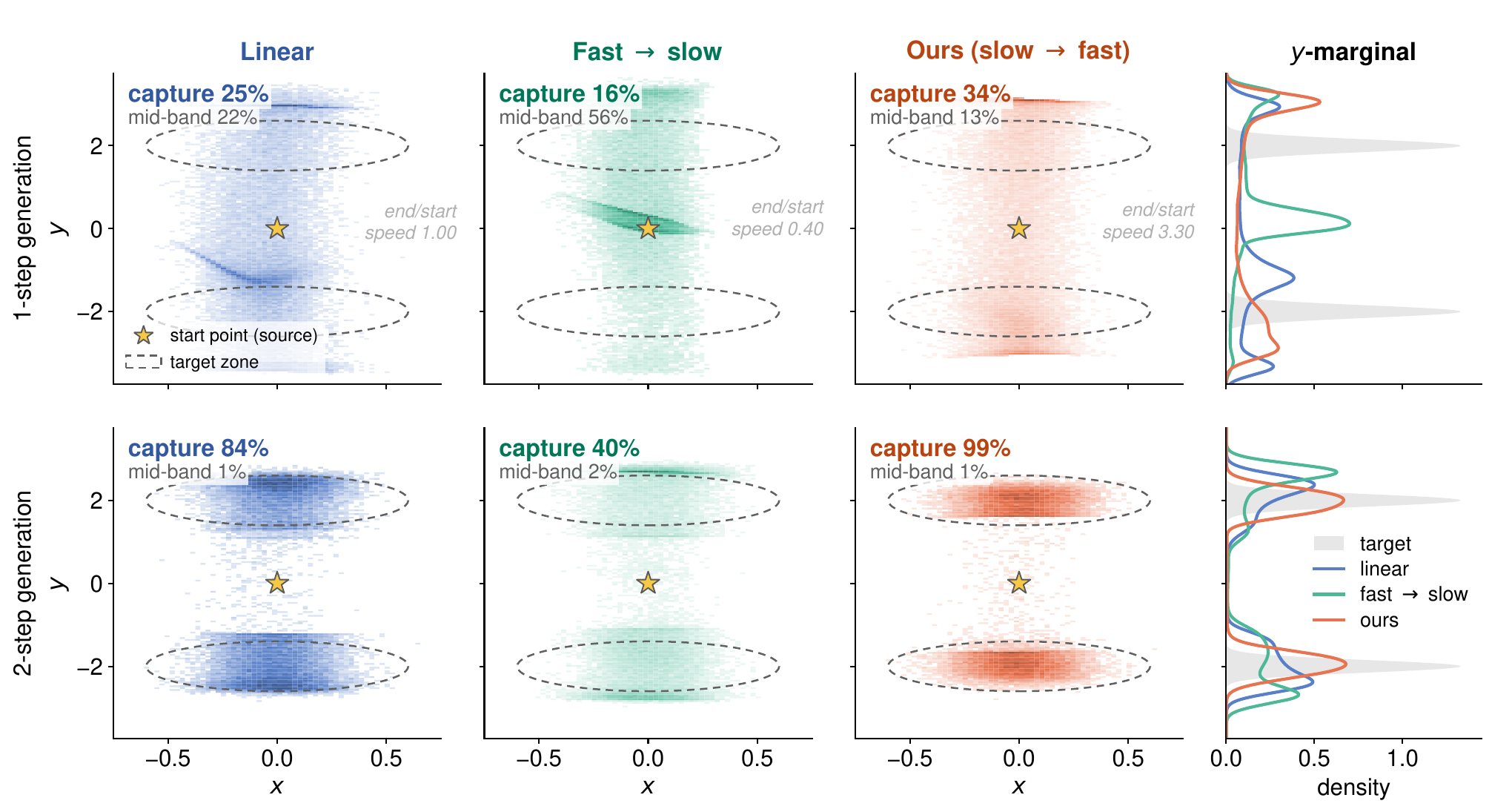}
    \caption{\textbf{Interpolation scheduling for few-step generation.}
The left three columns compare linear (standard), fast-then-slow,
and slow-then-fast (ours) scheduling, respectively.
The two rows show generated endpoints after one and two
integration steps, corresponding to NFE = 2 and 4, respectively. Stars mark the source mean, dashed outlines indicate the target
capture regions, and the rightmost column compares the generated
$y$-marginals with the target distribution (gray).
With two steps, capture reaches 84\%, 40\%, and 99\%, respectively.
With one step, fast-then-slow scheduling leaves 56\% of samples in
the central band, compared with 22\% for linear and 13\% for ours.
Slow-then-fast scheduling achieves the highest capture at both
integration budgets, supporting our scheduling design.}
    \label{fig:toy_exp}
\end{figure}

For completeness we report the full breakdown behind the two aggregate scores of the main
paper: the six reported LIBERO-Plus perturbation categories and every chain length on both CALVIN splits.
Both tables follow the method grouping and the marking convention of the main results table.

\paragraph{LIBERO-Plus: results by perturbation category.}
Tab.~\ref{tab:liberoplus_categories} reports results for six
perturbation categories, with each variant evaluated once under
each of three evaluation seeds. The weighted mean covers the
2,100 variants in these categories. We follow the six-category evaluation used in the main table,
excluding Camera.

For \m{}, the offline-trained language selector described in
App.~\ref{app:simulation_implementation} takes the provided instruction $\ell$ as input
and predicts $\hat{\tau}=\psi_\phi(\ell)$, which determines
the source $q_{\hat{\tau}}$ and its associated schedule
$\mathbf{K}_{\hat{\tau}}$.
In the Language instructions category, the perturbed instruction is
used for both source--schedule selection and policy conditioning;
benchmark task identities are not used for source--schedule selection.
This evaluates the complete language-conditioned inference pipeline
under instruction perturbations.
App.~\ref{app:open_language} provides a complementary evaluation of language-based task-identification accuracy on CALVIN.

\m{} achieves the highest weighted success rate (35.29\%),
followed by the fixed-$M_\tau$ source-only variant (35.21\%)
and MeanFlow (34.73\%).
Category-level rankings differ: MeanFlow obtains the highest
success rates on Robot initial states (76.1\%) and Language
instructions (72.7\%), compared with 73.6\% and 72.6\% for \m{}.
LeaP achieves 30.72\% weighted success and leads on Sensor
noise (12.2\%) and Lighting (8.9\%), while the fixed-$M_\tau$
source-only variant leads on Object and Background perturbations.
Relative to Standard FM, \m{} retains its largest gains on
Robot initial states (+19.6 percentage points) and Language
instructions (+18.6 percentage points).
The language-category result evaluates task execution with
the perturbed instruction driving both source--schedule selection
and policy conditioning.
Both source-only variants also improve weighted success over
Standard FM.

\paragraph{CALVIN: long-horizon chains on both splits.}
Tab.~\ref{tab:calvin_chain} reports 1,000 five-task chains
evaluated in environment D, with each chain terminated at
its first failure.
Column $\geq k$ gives the percentage of chains whose first
$k$ subtasks all succeed; summing these five percentages and
dividing by 100 recovers the average chain length, up to rounding.
\m{} achieves the highest mean chain length on both splits.
On D$\to$D, it also leads at every chain length, completing
all five subtasks in 19.3\% of chains, compared with 13.7\%
for MeanFlow and 10.0\% for LeaP.
On ABC$\to$D, \m{} reaches a mean chain length of 1.05,
compared with 0.86 for MeanFlow and 0.73 for LeaP.
It leads at $\geq1$ and $\geq2$, while PriorGrad achieves
the highest completion rates at $\geq3$, $\geq4$, and $\geq5$.
The five-task completion rate of \m{} is 1.7\%, compared
with 0.6\% for MeanFlow and 0.5\% for LeaP.

\subsubsection{Transfer to the Large-Scale \texorpdfstring{$\pi_{0.5}$}{pi0.5} Policy}
\label{app:pi05}

\paragraph{Implementation protocol.}
We initialize from the official \texttt{pi05\_base} checkpoint and fine-tune
the OpenPI JAX implementation using its standard low-memory LoRA recipe.
PaliGemma and the action expert use rank-16 and rank-32 LoRA adapters,
respectively, while both the baseline and \m{} otherwise retain the same
model architecture, optimization setup, data, and action normalization.
Each method is trained for a total of 14K optimization steps.

TOPS resumes the complete training state from the 6K-step baseline
checkpoint and continues training to 14K steps. From this checkpoint, we construct the
task-specific sources and schedules offline and continue training to the
same 14K-step budget, modifying only the source distribution and transport
targets. The baseline is evaluated from its corresponding 14K-step
checkpoint. At inference, the baseline uses ten Euler steps, whereas TOPS
uses one schedule-aware two-stage step (NFE=2). The $\pi_{0.5}$ action chunk contains ten 7-dimensional actions
($D=70$). We apply TOPS to these active action dimensions: expert chunks
are reverse-mapped through the intermediate baseline flow, followed by
per-task MPPCA source fitting and the same covariance-based scheduling
construction described in Section~\ref{sec:method}.

\paragraph{LIBERO-Plus perturbation breakdown.}
Tab.~\ref{tab:pi05_libero_plus} shows that the overall gain is distributed
across most perturbation categories. The largest improvements occur under
lighting changes (77.5\%$\rightarrow$94.2\%) and object-layout perturbations
(74.1\%$\rightarrow$84.3\%). TOPS also improves background, sensor-noise,
camera-viewpoint, and robot-initial-state robustness, while performance under
language perturbations remains comparable, with a small decrease from 74.8\%
to 73.7\%. Overall success increases from 64.4\% to 69.0\%.

\paragraph{CALVIN long-horizon breakdown.}
Tab.~\ref{tab:pi05_calvin} shows consistent gains across chain depths.
On D$\rightarrow$D, TOPS improves completion at every depth and raises
the average chain length from 3.673 to 3.941. The difference becomes more
pronounced under cross-environment transfer: on ABC$\rightarrow$D,
five-task completion increases from 41.2\% to 61.1\%, while the average
chain length rises from 3.033 to 3.744. The gains across deeper chain
positions suggest that the improvement is not confined to initial task
execution but persists over long-horizon rollouts.

\begin{table}[t]
\centering
\caption{
LIBERO-Plus perturbation breakdown for $\pi_{0.5}$ and $\pi_{0.5}$ + TOPS.
Success rates (\%) are reported for each perturbation category and overall.
}
\label{tab:pi05_libero_plus}
\small
\begin{tabular}{lcc}
\toprule
Perturbation & $\pi_{0.5}$ & $\pi_{0.5}$ + TOPS \\
\midrule
Lighting              & 77.5 & \textbf{94.2} \\
Object layout         & 74.1 & \textbf{84.3} \\
Background texture    & 79.0 & \textbf{81.9} \\
Sensor noise          & 46.8 & \textbf{51.8} \\
Camera viewpoint      & 38.4 & \textbf{40.0} \\
Robot initial state   & 74.4 & \textbf{76.1} \\
Language instruction  & \textbf{74.8} & 73.7 \\
\midrule
Overall               & 64.4 & \textbf{69.0} \\
\bottomrule
\end{tabular}
\end{table}

\begin{table}[t]
\centering
\caption{
CALVIN chain-depth results for $\pi_{0.5}$ and $\pi_{0.5}$ + TOPS.
Columns $\geq k$ report the percentage of five-task chains whose first
$k$ subtasks are completed successfully; Len denotes average completed
chain length.
}
\label{tab:pi05_calvin}
\small
\begin{tabular}{llcccccc}
\toprule
Setting & Method & $\geq1$ & $\geq2$ & $\geq3$ & $\geq4$ & $\geq5$ & Len \\
\midrule
D$\rightarrow$D
& $\pi_{0.5}$
& 87.4 & 79.2 & 72.3 & 66.7 & 61.7 & 3.673 \\
& $\pi_{0.5}$ + TOPS
& \textbf{90.7} & \textbf{83.7} & \textbf{78.4}
& \textbf{73.0} & \textbf{68.3} & \textbf{3.941} \\
\midrule
ABC$\rightarrow$D
& $\pi_{0.5}$
& 83.4 & 69.1 & 59.0 & 50.6 & 41.2 & 3.033 \\
& $\pi_{0.5}$ + TOPS
& \textbf{91.4} & \textbf{80.9} & \textbf{73.7}
& \textbf{67.3} & \textbf{61.1} & \textbf{3.744} \\
\bottomrule
\end{tabular}
\end{table}

\providecommand{\sd}[1]{{\scriptsize$\pm$#1}}
\begin{table*}[t]
\centering
\footnotesize
\setlength{\tabcolsep}{4pt}
\renewcommand{\arraystretch}{1.0}
\caption{
LIBERO-Plus results by perturbation category.
Success rates (\%), with variant counts in parentheses.
Type denotes the method category.
Weighted is the variant-count-weighted mean over the six
reported categories (Camera excluded, since all methods are zero).
Uncertainty follows the seed and aggregation convention of
Tab.~\ref{tab:main_results_compact}.
Bold and underlined values indicate the highest and
second-highest means in each column, respectively.
}
\label{tab:liberoplus_categories}
\begin{tabular}{@{}llrrrrrrr@{}}
\toprule
Method & Type
& Robot & Lang. & Object & Backgr. & Sensor & Light & Weighted \\
& & (393) & (383) & (312) & (289) & (449) & (274) & (2100) \\
\midrule
Standard FM & Ref.
& 54.0 & 54.0 & 21.7 & 13.6 & 4.7 & \underline{8.2}
& 27.1\sd{0.3} \\
OT-CFM & Coupl.
& 41.1 & 40.8 & 10.0 & 4.5 & 6.2 & 3.3
& 19.0\sd{0.1} \\
2-ReFlow & Few-step
& 38.1 & 42.0 & 17.4 & 9.6 & 4.5 & 4.4
& 20.2\sd{0.4} \\
RF-distill & Few-step
& 33.1 & 35.0 & 15.3 & 8.4 & 3.6 & 6.1
& 17.6\sd{0.1} \\
MeanFlow & Few-step
& \textbf{76.1} & \textbf{72.7} & 22.1 & 10.9
& \underline{8.4} & 4.9
& 34.7\sd{0.3} \\
PriorGrad & Source
& 62.3 & 61.7 & 23.2 & 14.5 & 7.5 & 4.6
& 30.6\sd{0.3} \\
MM-FM & Source
& 36.7 & 38.8 & 10.4 & 3.9 & 2.8 & 1.0
& 16.8\sd{0.5} \\
LeaP & Source
& 56.9 & 56.8 & \underline{26.1} & 15.1
& \textbf{12.2} & \textbf{8.9}
& 30.7\sd{0.8} \\
Regression & None
& 50.1 & 46.3 & 13.8 & 5.5 & 4.8 & 3.9
& 22.2\sd{0.5} \\
\midrule
Source (fixed $M_\tau$) & Ours
& 71.8 & 71.4 & \textbf{26.2} & \textbf{15.8} & 7.7 & 8.0
& \underline{35.2\sd{0.7}} \\
Source (task-wise $M_\tau$) & Ours
& 64.0 & 60.9 & 23.0 & 12.8 & 7.4 & 5.1
& 30.5\sd{0.7} \\
\textbf{\m{}} & Ours
& \underline{73.6} & \underline{72.6} & 24.8
& \underline{15.2} & 7.1 & 7.4
& \textbf{35.3\sd{0.1}} \\
\bottomrule
\end{tabular}
\end{table*}

\providecommand{\sd}[1]{{\scriptsize$\pm$#1}}
\begin{table*}[t]
\centering
\footnotesize
\setlength{\tabcolsep}{3pt}
\caption{\textbf{CALVIN long-horizon chains.} Average completed chain length and the
percentage of the 1000 chains whose first $k$ subtasks all succeeded ($\geq k$); higher is
better throughout. Type denotes the method category.}
\label{tab:calvin_chain}
\begin{tabular}{@{}ll cccccc cccccc@{}}
\toprule
& & \multicolumn{6}{c}{D$\to$D (in-distribution)} & \multicolumn{6}{c}{ABC$\to$D (unseen env.)} \\
\cmidrule(lr){3-8}\cmidrule(lr){9-14}
Method & Type & Len & $\geq$1 & $\geq$2 & $\geq$3 & $\geq$4 & $\geq$5
              & Len & $\geq$1 & $\geq$2 & $\geq$3 & $\geq$4 & $\geq$5 \\
\midrule
Standard FM & Ref.   & 1.90\sd{.05} & 72.9 & 52.2 & 32.2 & 20.3 & 11.9
                     & 0.70\sd{.03} & 42.8 & 16.8 & 7.0 & 2.5 & 0.5 \\
OT-CFM      & Coupl. & 1.28\sd{.05} & 59.8 & 34.6 & 18.8 & 9.8 & 4.9
                     & 0.46\sd{.02} & 32.9 & 10.0 & 2.8 & 0.6 & 0.1 \\
2-ReFlow    & Few-step & 1.74\sd{.05} & 72.0 & 47.5 & 28.6 & 17.1 & 8.9
                     & 0.66\sd{.03} & 42.3 & 15.8 & 5.5 & 2.4 & 0.4 \\
RF-distill  & Few-step & 1.75\sd{.05} & 71.0 & 48.3 & 28.9 & 17.0 & 9.9
                     & 0.70\sd{.03} & 44.1 & 17.7 & 6.5 & 1.7 & 0.2 \\
MeanFlow & Few-step
& 1.89\sd{.02} & 71.4 & 49.2 & 32.9 & 22.0 & 13.7
& 0.86\sd{.02} & 50.5 & 23.9 & 8.6 & 2.7 & 0.6 \\
PriorGrad   & Source & 1.97\sd{.06} & 72.4 & 50.7 & 35.4 & 23.3 & 15.0
                     & \underline{1.02\sd{.04}} & \underline{52.5} & \underline{27.4}
                     & \textbf{12.9} & \textbf{6.4} & \textbf{2.8} \\
MM-FM       & Source & 1.75\sd{.05} & 70.3 & 47.9 & 29.7 & 17.9 & 9.0
                     & 0.68\sd{.03} & 42.4 & 16.4 & 6.6 & 2.1 & 0.2 \\
LeaP & Source
& 1.74\sd{.02} & 70.1 & 46.9 & 29.0 & 17.7 & 10.0
& 0.73\sd{.03} & 45.8 & 18.2 & 7.1 & 1.9 & 0.5 \\
Regression  & None   & 0.88\sd{.03} & 53.0 & 21.3 & 8.9 & 4.0 & 1.0
                     & 0.27\sd{.02} & 21.8 & 4.2 & 0.6 & 0.0 & 0.0 \\
\midrule
Source (fixed $M_\tau$) & Ours
  & 2.12\sd{.06} & 74.4 & 54.7 & \underline{39.2} & \underline{26.8} & 17.3
  & 0.88\sd{.04} & 49.7 & 23.4 & 9.6 & 4.2 & 1.4 \\
Source (task-wise $M_\tau$) & Ours
  & \underline{2.15\sd{.06}} & \underline{76.0} & \underline{55.6} & \underline{39.2}
  & \underline{26.8} & \underline{17.6}
  & 0.98\sd{.04} & 52.1 & 26.2 & \underline{12.1} & \underline{6.0} & \underline{1.7} \\
\textbf{\m{}} & Ours
  & \textbf{2.27\sd{.06}} & \textbf{77.0} & \textbf{57.9} & \textbf{42.5} & \textbf{30.5}
  & \textbf{19.3}
  & \textbf{1.05\sd{.04}} & \textbf{56.8} & \textbf{29.0} & 12.2 & 5.5 & \underline{1.7} \\
\bottomrule
\end{tabular}
\end{table*}

\subsection{Language-based Source Selection}
\label{app:language_selection}

\m{} separates source selection from the policy's conditioning pathway.
Task annotations organize the offline construction of the source
distributions $\{q_\tau\}_{\tau\in\mathcal T}$ and their associated
scheduling matrices $\{K_\tau\}_{\tau\in\mathcal T}$.
At inference, a learned selector $\psi_\phi$ maps the input instruction
$\ell$ to a source index $\hat\tau\in\mathcal T$:
\begin{equation}
\label{eq:language_source_interface}
    \hat\tau=\psi_\phi(\ell),
    \qquad
    x^{(0)}\sim q_{\hat\tau}.
\end{equation}
The selected source is used with its corresponding fixed scheduling
matrix $K_{\hat\tau}$.

In the main-text evaluations, the learned selector operates
separately from the velocity network. It uses MiniLM~\citep{minilm} sentence
features to predict the task index, while the policy condition $\textbf{c}$
is constructed from observations and the provided instruction
through the original observation encoders and frozen CLIP text
encoder, independently of that prediction.
Action generation then integrates $v_\theta(\mathbf{x},t,\textbf{c})$ from the
selected source using its associated schedule.
The selector implementation is provided in App.~\ref{app:simulation_implementation}. App.~\ref{app:open_language} compares language-based task-identification methods on 358 instruction formulations covering 34 known CALVIN tasks.

\subsection{Empirical Characterization of Expansion Directions}
\label{app:expansion-analysis}

\subsubsection{Spectral Geometry of Expansion Directions}
\label{app:expansion-spectral-geometry}
Fig.~\ref{fig:expansion-diagnostics} examines the covariance geometry
motivating the direction-dependent interpolation schedules in Section~\ref{sec:raft_RAFT}:
the concentration of expansion directions, the stability of their estimated
subspaces, and their sensitivity to the source covariance.

\paragraph{Diagnostic setup.}
We analyze LIBERO-10 and the CALVIN D and ABC training settings,
with 10, 34, and 34 task-setting pairs, respectively; the CALVIN settings
share task categories.
These offline diagnostics use fixed three-component, rank-20 MPPCA sources
with source scale $0.3$ and policy-normalized 16-step action chunks
($D=16\times10=160$).
Source covariances are computed analytically from $q_\tau$.
Expert covariances are estimated from stride-one windows within stored
episodes, with $5\%$ shrinkage toward trace-matched isotropic covariances.
We follow the OT construction in Section~\ref{sec:raft_RAFT}, using its eigenvalues
$s_{\tau j}$ and unit eigenvectors $\boldsymbol r_{\tau j}\in\mathbb R^D$.
All reported averages weight tasks equally.

\paragraph{Expansion and concentration.}
Define the expansion set
$\mathcal E_\tau=\{j:s_{\tau j}>1\}\subseteq\{1,\ldots,D\}$ and its cardinality
$k_\tau=|\mathcal E_\tau|\in\{0,\ldots,D\}$.
Define the scalar covariance-cost contribution
$e_{\tau j}\in\mathbb R_{\geq0}$ and expansion cost share
$R_{\mathrm{exp}}^\tau\in[0,1]$:
\begin{equation}
    e_{\tau j}
    =(s_{\tau j}-1)^2
    \boldsymbol r_{\tau j}^{\top}\mathbf C_s^\tau\boldsymbol r_{\tau j},
    \qquad
    R_{\mathrm{exp}}^\tau
    =\frac{\sum_{j\in\mathcal E_\tau}e_{\tau j}}
           {\sum_{j=1}^{D}e_{\tau j}}.
    \label{eq:expansion-diagnostic-cost}
\end{equation}
The denominator is the squared Bures distance between the two covariances,
excluding mean displacement.
Fig.~\ref{fig:expansion-diagnostics}(a,b) shows that expansion is concentrated
in a small subspace: the mean expansion fractions are
$7.4\%$, $9.2\%$, and $9.7\%$ for LIBERO-10, CALVIN D, and CALVIN ABC,
while their mean covariance-cost shares are $69.6\%$, $60.9\%$, and $62.0\%$.
Thus, distinguishing expansion directions targets a substantial portion of
the observed second-order mismatch, providing empirical motivation for the
direction selection in Section~\ref{sec:raft_RAFT}.

\paragraph{Stability under expert-data resampling.}
For column-orthonormal bases
$\mathbf Q_A\in\mathbb R^{D\times k_A}$ and
$\mathbf Q_B\in\mathbb R^{D\times k_B}$, with
$k_A,k_B\in\{1,\ldots,D\}$, define the symmetric subspace overlap
\begin{equation}
    \mathcal O(\mathbf Q_A,\mathbf Q_B)
    =\frac{1}{2}\left(\frac{1}{k_A}+\frac{1}{k_B}\right)
      \|\mathbf Q_A^{\top}\mathbf Q_B\|_F^2
    \in[0,1].
    \label{eq:expansion-diagnostic-overlap}
\end{equation}
Its expectation for independent uniformly random subspaces is
$(k_A+k_B)/(2D)$.
We perform 20 random splits of stored episode IDs, re-estimate the expert
covariance in each half, and compare the resulting top-$k_\tau$ OT subspaces.
The source covariance, action normalization, and full-data choice of
$k_\tau$ remain fixed.
The task-mean overlaps in Fig.~\ref{fig:expansion-diagnostics}(c) are
$0.993$, $0.973$, and $0.984$, compared with random-reference means of
$0.074$, $0.092$, and $0.097$.
This supports stable offline subspace estimation under the specified protocol.
Splits group complete demonstration episodes for LIBERO and annotated language
segments for CALVIN; the latter need not be independent play trajectories.

\paragraph{Sensitivity to the source covariance.}
Holding $\mathbf C_d^\tau$ fixed, we replace the fitted source covariance
with $0.3^2\mathbf I_D$ and recompute the OT map.
We compare each source's own expansion subspace using
Eq.~\ref{eq:expansion-diagnostic-overlap}, allowing their dimensions to differ.
Fig.~\ref{fig:expansion-diagnostics}(d) gives mean overlaps of
$0.877$, $0.811$, and $0.780$, above the corresponding random-reference means
of $0.069$, $0.079$, and $0.084$.
Thus, substantial expansion structure persists for both source choices,
supporting the premise in Section~\ref{sec:raft_RAFT} that source adaptation can leave
covariance mismatch that motivates direction-dependent interpolation scheduling.

\subsubsection{Action Structure of Expansion Directions}
\label{app:expansion-action-structure}

\begin{figure}[t]
    \centering
    \includegraphics[width=0.8\linewidth]{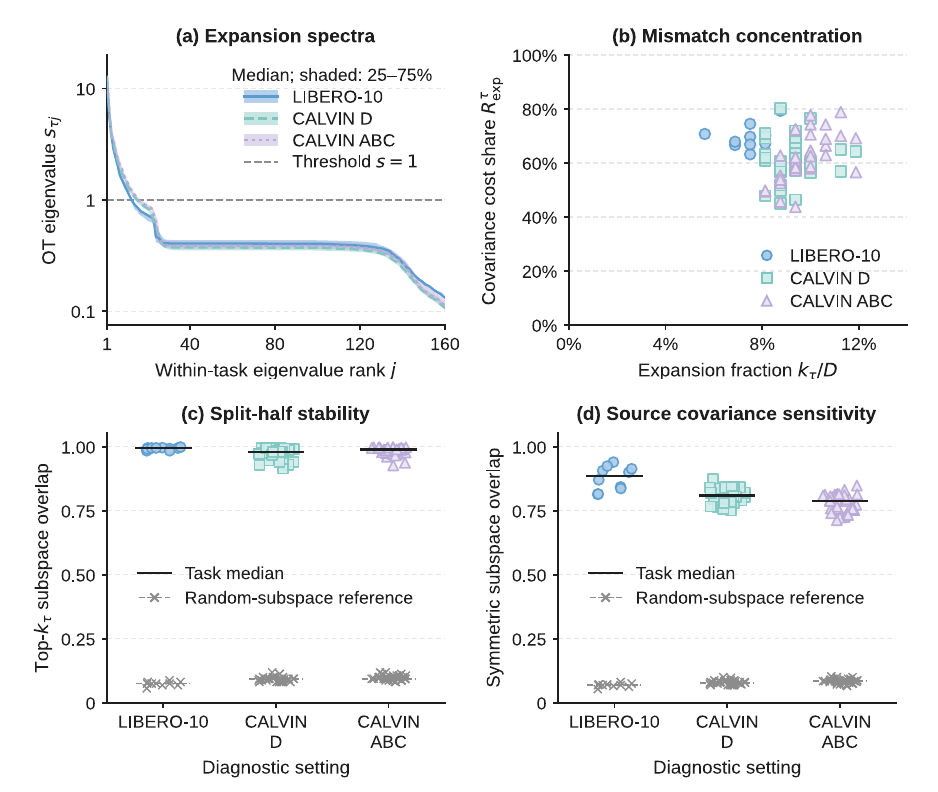}
    \caption{
    \textbf{Empirical characterization of expansion directions.}
    (a) Within-task sorted OT eigenvalues: curves show task medians and bands
    show interquartile ranges; the dashed line marks $s_{\tau j}=1$.
    (b) Expansion fraction $k_\tau/D$ versus covariance-cost share
    $R_{\mathrm{exp}}^\tau$.
    (c) Top-$k_\tau$ subspace overlap, averaged over 20 random data splits
    with the source and full-data dimension fixed.
    (d) Symmetric overlap between the fitted-source and isotropic-source
    expansion subspaces, each selected by $s_{\tau j}>1$.
    Each point in (b--d) represents a task within a diagnostic setting.
    Black bars in (c,d) show task medians; grey markers show analytical
    random-subspace expectations.
    All spectra use the regularized covariance estimates specified above.
    }
    \label{fig:expansion-diagnostics}
\end{figure}

Section~\ref{sec:raft_RAFT} interprets the OT eigenvectors
$\boldsymbol r_{\tau j}\in\mathbb R^D$ as action patterns spanning
timesteps and action dimensions.
Complementing the spectral diagnostics in
Fig.~\ref{fig:expansion-diagnostics}, we examine where these directions
load in action coordinates.
We use the same LIBERO-10 diagnostic setup as
App.~\ref{app:expansion-analysis}, with $H=16$, $d_a=10$, and
$D=Hd_a=160$ in policy-normalized action space.

\paragraph{Coordinate-loading diagnostic.}
Let $\mathcal E_\tau^c=\{1,\ldots,D\}\setminus\mathcal E_\tau$
denote the non-expansion set.
Using the covariance-cost contributions $e_{\tau j}$ from
Eq.~\ref{eq:expansion-diagnostic-cost}, define the nonnegative loading
matrix $\mathbf L_\tau(\mathcal S)\in\mathbb R_{\geq0}^{H\times d_a}$
for $\mathcal S\in\{\mathcal E_\tau,\mathcal E_\tau^c\}$ with
$\sum_{j\in\mathcal S}e_{\tau j}>0$:
\begin{equation}
    [\mathbf L_\tau(\mathcal S)]_{h,a}
    =
    \frac{
        \sum_{j\in\mathcal S}e_{\tau j}
        [\boldsymbol r_{\tau j}]_{(h-1)d_a+a}^{\,2}
    }{
        \sum_{j\in\mathcal S}e_{\tau j}
    },
    \qquad
    h\in\{1,\ldots,H\},\quad a\in\{1,\ldots,d_a\}.
    \label{eq:expansion-coordinate-loadings}
\end{equation}
Because the eigenvectors have unit norm, each matrix satisfies
$\sum_{h=1}^{H}\sum_{a=1}^{d_a}
[\mathbf L_\tau(\mathcal S)]_{h,a}=1$.
We average these matrices equally over the ten tasks.
Group shares sum the resulting entries over all timesteps and coordinates
$\{1,2,3\}$ for position, $\{4,\ldots,9\}$ for the 6D rotation
representation, and $\{10\}$ for the gripper.
Uniform loading over coordinates gives reference shares of
$30\%$, $60\%$, and $10\%$, respectively.

\begin{figure}[t]
    \centering
    \includegraphics[width=\linewidth]{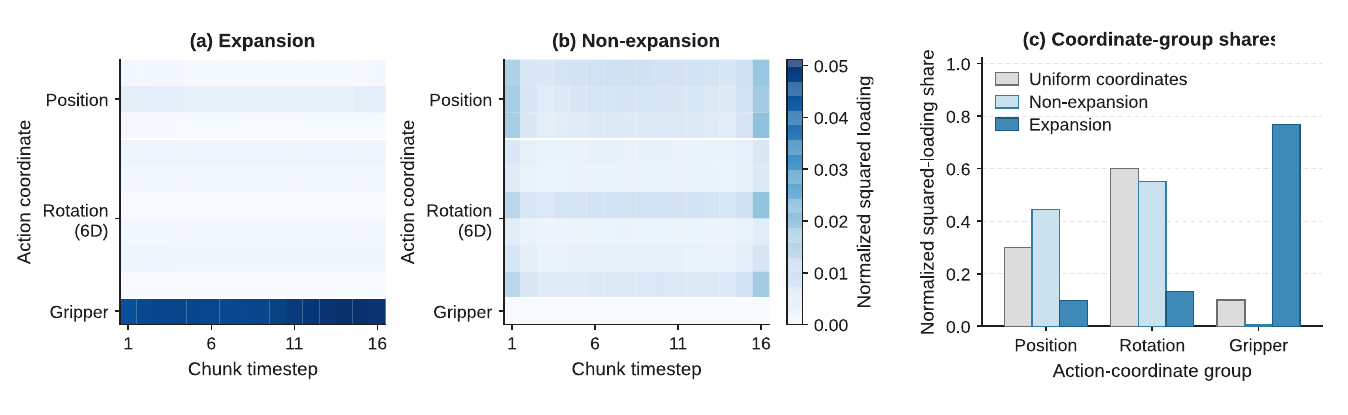}
    \vspace{-20pt}
    \caption{
    \textbf{Action-coordinate structure of OT directions on LIBERO-10.}
    (a,b): covariance-cost-weighted squared loadings
    for expansion ($s_{\tau j}>1$) and non-expansion ($s_{\tau j}\leq1$)
    directions, respectively, defined in
    Eq.~\ref{eq:expansion-coordinate-loadings}.
    Each task's map is independently normalized to sum to one before
    averaging across the ten tasks; both heatmaps use the same color scale.
    Columns index chunk timesteps and rows index action coordinates.
    (c): loading shares summed over position, rotation, and
    gripper coordinates. Grey bars indicate the uniform-coordinate
    reference determined by group dimensionality.
    The figure analyzes raw \textit{Gaussian}-OT eigenvectors in normalized
    action space.
    }
    \label{fig:expansion-loadings}
\end{figure}

\paragraph{Results and connection to Section~\ref{sec:raft_RAFT}.}
Fig.~\ref{fig:expansion-loadings} reveals distinct action-coordinate
profiles for expansion and non-expansion directions.
Gripper coordinates account for $76.9\%$ of the normalized expansion
loading, compared with the $10\%$ uniform-coordinate reference;
the averaged expansion map shows gripper participation throughout the
16-step horizon.
Non-expansion loadings instead concentrate on position ($44.4\%$)
and rotation ($55.1\%$), with only $0.5\%$ on the gripper.
This action-group selectivity provides empirical support for the
action-structure interpretation of the OT directions in Section~\ref{sec:raft_RAFT},
complementing the preceding evidence about their spectral geometry.
The diagnostic characterizes aggregate coordinate participation:
averaging squared loadings does not retain the signed temporal
coordination of individual modes, and these shares are not
coordinate-wise transport-cost contributions.

\subsection{Ablation of the schedule-aware integrator}
\label{app:solver_ablation}

\paragraph{Experimental setup.}
We evaluate the effect of the inference solver on LIBERO-10 while keeping
the TOPS checkpoint, task-specific source distributions $q_\tau$, and
learned velocity field $v_\theta$ fixed, without retraining.
This diagnostic uses $M_\tau=3$ source components per task, component rank
$r=20$, and source scale $0.3$.
We retain a prediction horizon of 16, two observation frames, and eight
executed actions per policy call.
For each solver and budget, we evaluate EMA weights from one
continuation-training seed, using 20 rollouts per task, fixed initial
states, a maximum of 400 environment steps, and three evaluation seeds.
We report the equally weighted ten-task success rate as mean $\pm$
population standard deviation across three evaluation seeds for one
fixed trained checkpoint, totaling 600 rollouts per table entry.

\paragraph{Solvers and evaluation budgets.}
We compare the two-stage schedule-aware update in App.~\ref{app:integrator_details} with
explicit midpoint and ordinary explicit Euler integration of the same
learned field.
On a uniform grid $t_n=n/N$ with $h=1/N$, Euler uses
\[
  x^{(n+1)}=x^{(n)}+h\,v_\theta(x^{(n)},t_n,c).
\]
Euler requires one velocity-network evaluation per step, whereas explicit
midpoint and the schedule-aware update require two.
We therefore compare NFE budgets of $2$, $10$, and $20$: Euler uses
$2$, $10$, and $20$ steps, while the other solvers use $1$, $5$, and $10$.
The schedule-aware integrator applies two successive half-interval
updates defined in App.~\ref{app:integrator_details}. It differs from standard explicit
midpoint and reduces to two explicit Euler half-steps when
$\mathbf{K}_{\tau}=0$.

\begin{table}[t]
\centering
\small
\setlength{\tabcolsep}{6pt}
\caption{\textbf{Solver ablation on LIBERO-10 with a fixed TOPS checkpoint.}
Success rate (\%, higher is better), reported as mean $\pm$ population
standard deviation over three evaluation seeds.
All rows use the same $M_\tau=3$ checkpoint and differ only in the
inference solver. Columns align the number of velocity-network evaluations
(NFE), rather than the number of integration steps.}
\label{tab:solver_ablation_libero10}
\begin{tabular}{@{}lccc@{}}
\toprule
Integrator & NFE $=2$ & NFE $=10$ & NFE $=20$ \\
\midrule
Explicit Euler
& $0.00\pm0.00$ & $35.67\pm1.43$ & $55.00\pm2.16$ \\
Explicit midpoint
& $0.00\pm0.00$ & $57.67\pm1.84$ & $57.50\pm2.48$ \\
Schedule-aware (ours)
& $\mathbf{68.33\pm2.09}$ & $\mathbf{72.17\pm4.48}$
& $\mathbf{72.67\pm2.01}$ \\
\bottomrule
\end{tabular}
\end{table}

\paragraph{Results.}
Tab.~\ref{tab:solver_ablation_libero10} shows that the schedule-aware
integrator achieves the highest mean success rate at every tested NFE
budget. At NFE~$=2$, it reaches $68.33\%$, while neither generic solver
achieves a successful rollout. Even at this budget, it outperforms both
generic solvers evaluated at NFE~$=20$.
Increasing its budget to NFE~$=10$ and $20$ further raises mean success
to $72.17\%$ and $72.67\%$, respectively.
At these two budgets, the margins over explicit Euler are $36.50$ and
$17.67$ percentage points, while those over explicit midpoint are
$14.50$ and $15.17$ points.
These results support the effectiveness of the schedule-aware update
for few-step inference, with additional evaluations providing further
improvements in mean task success.

\subsection{Implementation details}

\subsubsection{Implementation Details for Simulation Experiments}
\label{app:simulation_implementation}

\paragraph{Policy architecture.}
All methods share a diffusion Transformer (DiT) backbone with four blocks,
a hidden dimension of 512, and eight attention heads. A ResNet-18 encoder
extracts visual features, while a frozen CLIP ViT-B/32 encoder provides
512-dimensional language embeddings.
These features are combined with proprioceptive observations to condition
the DiT through adaptive layer normalization. The network predicts
velocities for action chunks, and TOPS preserves this shared architecture.
We infer the task index from the instruction using a frozen
MiniLM sentence encoder~\citep{minilm,sentencebert} followed by
a five-model ensemble of lightweight MLP classifiers.
The classifiers are trained offline on the benchmark-provided
training instructions and their task labels using a standard
cross-entropy objective and AdamW with a learning rate of $10^{-3}$,
independently of the policy.
The sentence encoder remains frozen, and the trained selector
is held fixed during policy training and evaluation.
For \m{}, its predicted task index selects both $q_{\hat{\tau}}$
and $K_{\hat{\tau}}$; the source-only variants use the same
prediction to select $q_{\hat{\tau}}$.
The policy retains its original CLIP-based language conditioning.

\paragraph{Training and hyperparameters.}
We first train the standard FM policy for 300 warm-up epochs. We then
construct the task-specific sources $q_\tau$ and scheduling matrices
$\mathbf{K}_\tau$ offline and continue training for another 300 epochs,
keeping both fixed. \textbf{All compared methods are trained for the same
total of 600 epochs, including any warm-up stages, to ensure a fair
comparison.} Each epoch comprises 250 optimizer updates. We evaluate the
final exponential moving average (EMA) policy. The main hyperparameters
are summarized in Tab.~\ref{tab:simulation_hyperparameters}.

\paragraph{Random seeds and statistical reporting.}
We distinguish training seeds from evaluation seeds.
An independent training run repeats the stochastic training
pipeline; for \m{}, this includes warm-up, source and schedule
construction, and continuation training.
An evaluation seed controls rollout stochasticity for a fixed
trained checkpoint.

We first compute the benchmark metric separately for each
evaluation seed. When multiple independently trained checkpoints
are evaluated, we average evaluation repeats within each
checkpoint before computing statistics across training runs.
For fixed-checkpoint experiments, statistics are computed across
evaluation seeds.
Each experiment specifies the corresponding level of repetition.
Unless otherwise specified, reported standard deviations are sample standard deviations,
computed with denominator $N-1$ for $N$ repeated measurements.

\paragraph{Method hyperparameters.}
For each task source, $M_\tau$ specifies the number of mixture components
and $r$ the dimension of their low-rank covariance term. We calibrate the
pooled source standard deviation to the warm-up \textit{Gaussian} scale,
$\sqrt{\operatorname{tr}(\mathbf{C}_{\mathrm s}^\tau)/D}=\sigma_0$.
For interpolation scheduling, $\kappa$ bounds the strengths $\rho_{\tau j}$,
while the sensitivity $\gamma$ is computed as
\begin{equation}
\gamma =
\left[
\max\!\left\{
\operatorname{median}_{\tau,j:\,s_{\tau j}>1}
\log s_{\tau j},\;10^{-6}
\right\}
\right]^{-1}.
\label{eq:simulation_gamma}
\end{equation}
The median pools the positive log expansion ratios of the \textit{Gaussian} OT
maps across training tasks. We compute $\gamma$ offline for each training
setting and keep it fixed during continuation training and inference.

\paragraph{Inference-time measurement.}
All inference latencies in Tab.~\ref{tab:main_results_compact}
are measured on an NVIDIA RTX PRO 6000 GPU.
We time one policy call that generates an action chunk,
using each method's reported inference configuration.

\begin{table}[ht]
\centering
\caption{Main hyperparameters for simulation experiments.}
\label{tab:simulation_hyperparameters}
\small
\setlength{\tabcolsep}{6pt}
\renewcommand{\arraystretch}{1.05}
\begin{tabular}{@{}ll@{}}
\toprule
Hyperparameter & Value \\
\midrule
TOPS warm-up / continuation epochs & $300\,/\,300$ \\
Total training epochs, all methods & 600 \\
Batch size & 32 \\
Optimizer & AdamW, $\beta_1=0.95$, $\beta_2=0.999$ \\
Learning rate & $10^{-4}$ \\
Weight decay & $10^{-6}$ \\
Learning-rate schedule & Cosine, with 500 warm-up updates per stage \\
Maximum EMA decay & 0.9999 \\
\midrule
Prediction / observation / execution horizons & $16\,/\,2\,/\,8$ \\
Mixture components $M_\tau$ & 2 \\
Low-rank dimension $r$ & 20 \\
Warm-up source standard deviation $\sigma_0$ & 0.3 \\
Calibrated pooled source standard deviation & 0.3 \\
Maximum scheduling strength $\kappa$ & 0.5 \\
Scheduling sensitivity $\gamma$ & Computed using Eq.~\ref{eq:simulation_gamma} \\
Reverse-flow integration & 10 explicit-midpoint steps \\
TOPS inference steps $N$ & 1 (two network evaluations) \\
\bottomrule
\end{tabular}
\end{table}

\paragraph{Baseline implementation protocol.}
We compare baseline adaptations within a shared policy framework,
using the same observation encoder, language conditioning, DiT
backbone, action representation, and observation and execution
horizons.
Each normalized action chunk contains 16 ten-dimensional actions
($D=160$).
All methods use the same optimizer settings and receive 600 epochs
of policy training, including any warm-up stage.
Continuation-based methods initialize from the EMA parameters of
a shared 300-epoch Standard FM checkpoint and train for another
300 epochs.
Offline prior fitting and teacher-pair generation are performed
separately before continuation.
Unless otherwise specified, inference uses ten explicit-midpoint
steps, corresponding to 20 velocity-network evaluations
(NFE $=20$).

\paragraph{Standard FM.}
We independently sample Gaussian sources
$x_0\sim\mathcal{N}(0,0.3^2I_D)$ and expert action chunks $x_1$.
Training uses linear interpolation
$x_t=(1-t)x_0+t x_1$, with $t\sim\mathcal{U}[0,1]$,
and mean-squared regression of the displacement $x_1-x_0$.
The same source distribution and objective are used throughout
training.

\paragraph{OT-CFM.}
We replace independent endpoint pairing with minibatch optimal
transport based on squared Euclidean distances in the flattened
action-chunk space.
The uniform-marginal transport problem is solved using exact
earth mover's distance, with Hungarian assignment as a fallback.
Matching operates across the full minibatch using action-space
costs.
Only source samples are reassigned; expert chunks retain their
associated observations and language instructions.
The remaining training procedure follows Standard FM.

\paragraph{2-ReFlow.}
We use the frozen warm-up EMA policy to generate four
noise--endpoint pairs per demonstration window, conditioned on
its observations and language instruction.
Teacher generation uses 50 explicit-midpoint steps (NFE $=100$).
During continuation, one cached pair is randomly selected per
window, and the student learns the linear FM objective between
the paired endpoints.
Inference uses one explicit-midpoint step (NFE $=2$).

\paragraph{RF-distill.}
We use the same cached teacher pairs as 2-ReFlow and train the
network to predict their complete displacement at $t=0$.
A fresh Gaussian sample is mapped to an action chunk through
$\hat{x}_1=z+v_\theta(z,0,c)$, requiring one network evaluation.
RF-distill directly distills the warm-up teacher and is trained
as a separate continuation branch from 2-ReFlow, within the
same 600-epoch policy-training budget.

\paragraph{MeanFlow.}
We augment the shared DiT with an endpoint-time embedding branch
to predict interval-average velocities.
The additional branch is initialized to preserve the warm-up
network's predictions.
Training uses the source-to-data form of the MeanFlow identity,
with stop-gradient targets computed through Jacobian--vector
products.
We use time-reversed logit-normal sampling with parameters
$\mu=-0.4$ and $\sigma=1.0$, and set the two interval endpoints
equal for 75\% of examples.
Detached inverse-error weights are normalized within each minibatch.
Both the main and limited-data experiments use one full-interval
update (NFE $=1$), without classifier-free guidance.

\paragraph{PriorGrad.}
We adapt the data-dependent prior to flow matching by fitting a
diagonal Gaussian to each task's normalized training actions.
The ten-dimensional mean and standard deviations are shared
across the action horizon, with independent samples at each
position.
Standard deviations are floored at $0.01$.
Training uses linear interpolation and inverse-variance-weighted
velocity regression.
The weights are clipped at their global 99th percentile and
normalized to mean one.
The fitted prior remains fixed during policy training and inference.

\paragraph{MM-FM source adaptation.}
For the MM-FM entry, we fit a task-specific diagonal Bayesian
Gaussian mixture to normalized expert action chunks.
We use up to eight components with a Dirichlet-process weight
prior and at most 40,000 unpadded training windows per task.
Components with weights below $0.01$ are removed, the remaining
weights are renormalized, and standard deviations are floored
at $0.01$.
The fitted source remains fixed during policy training.
This adaptation evaluates the mixture-source component using
independent source--target pairing within each task, in place
of the original mode-dependent coupling.
Training otherwise uses the standard linear FM objective.

\paragraph{LeaP.}
We learn a proprioception-conditioned diagonal Gaussian source
with a $16\rightarrow512\rightarrow20$ MLP.
Its input comprises two normalized proprioceptive observations,
and its outputs specify ten-dimensional means and log variances
shared across the action horizon.
Visual and language features condition the velocity network.
The source is initialized to the warm-up Gaussian and optimized
jointly with the policy through reparameterized sampling.
The training objective combines flow matching, Gaussian
negative log-likelihood, and symmetric source--action contrastive
alignment, with respective weights $1.0$, $1.0$, and $0.3$.
Alignment uses 128-dimensional projections and temperature $0.07$.
Our adaptation uses the shared image-based backbone and normalized
raw proprioception, with explicit-midpoint integration.
The main experiments use ten steps (NFE $=20$), and limited-data
experiments use five (NFE $=10$).
NFE counts velocity-network evaluations and excludes the prior
MLP forward pass.

\paragraph{Deterministic action-chunk regression.}
We use the shared backbone with a zero action input and fixed
time $t=0$ to predict the normalized expert chunk directly
with an L1 loss.
The predictor uses neither a variational latent model nor
temporal ensembling.
Inference requires one network evaluation, followed by the
same action unnormalization and eight-action execution used
by the other policies.

\paragraph{Comparison scope.}
The reported results compare these specified adaptations under
a shared policy framework, a common policy-training epoch budget,
and the stated inference budgets.
For teacher-based methods, continuation fits cached teacher
outputs, so performance depends jointly on the warm-up teacher
and the student's approximation at its inference budget.

\subsubsection{Implementation Details for Real-World Experiments}
\label{app:real-world-details}

\paragraph{Robot platform.} All real-world experiments are conducted on a Galaxea R1 Lite, a
wheeled dual-arm mobile manipulator whose upper body consists of two 6-DoF arms with parallel
two-finger grippers, an articulated torso, and a head unit carrying RGB cameras. We restrict
control to the right arm and its gripper; the left arm, the torso, and the mobile base are held
in a fixed configuration throughout both demonstration collection and evaluation, so that the
head-camera viewpoint is identical in the two phases.

\paragraph{Observation and action spaces.} Policies receive two RGB streams, one from the
head-mounted camera and one from the right wrist-mounted camera. Each frame is padded to a square
aspect ratio and resized to $256 \times 256$. Proprioception is a 7-dimensional vector containing
the six right-arm joint positions and the gripper opening. The language instruction of each task
is encoded offline by the frozen CLIP ViT-B/32 text encoder into a 512-dimensional embedding,
which is projected and concatenated with the visual and proprioceptive features to form the
conditioning vector of the DiT backbone. Actions are 7-dimensional targets comprising absolute
positions for the same six joints and a gripper command. We use the same prediction, observation,
and execution horizons as in simulation ($16 / 2 / 8$).

\paragraph{Demonstration collection.} Demonstrations are collected by human teleoperation and
recorded at $15$\,Hz, with camera frames and proprioceptive states synchronized to a common clock.
Within each suite, all candidate targets are present in a shared scene layout, and object
placements are varied across demonstrations within a fixed workspace region; consequently the
language instruction is the only signal that identifies the intended task at the start of an
episode. Every demonstration is annotated with the instruction it executes.

\paragraph{Training.} Real-world policies follow the same two-stage recipe and the same
optimization settings as in simulation (Tab.~\ref{tab:simulation_hyperparameters}). We first train the standard
FM policy for $500$ warm-up epochs. We then construct the task-specific sources $q_\tau$ and the
scheduling matrices $\mathbf{K_\tau}$ offline from this frozen warm-up policy, and continue training for
another $500$ epochs with both held fixed. All compared methods are trained for the same total of
$1000$ epochs, including any warm-up stage, and we deploy the final exponential moving average
(EMA) policy. Source and schedule construction reuses the simulation procedure: expert action
chunks are mapped backward through the warm-up flow with $10$ explicit-midpoint steps to form the
per-task reverse-image banks, each bank is fit with $M_\tau$ mixture components carrying a
rank-$r$ covariance term, the pooled source standard deviation is calibrated to $\sigma_0$, and
the scheduling strengths are bounded by $\kappa$ with the sensitivity $\gamma$ computed offline
per suite according to Eq.~\ref{eq:simulation_gamma}. The resulting values are listed in
Tab.~\ref{tab:real-hparams}.

\paragraph{Evaluation protocol.} Each task is evaluated over $20$ rollouts. A rollout is scored as a binary success: for Drawer Opening,
the instructed drawer is pulled open; for Color Matching, the instructed cube is released inside
the tray of the matching color; for Pick-Vegetable, the instructed object is released inside the
collection box. A rollout that does not satisfy its criterion within a fixed time budget is
counted as a failure. All methods are evaluated under the same object layouts and the same reset
procedure.

\paragraph{Aggregate statistical analysis.}
We compare \m{} and Standard FM by pooling binary outcomes
across the eight tasks, with 20 rollouts per task and method.
Each task therefore receives equal weight in the pooled
success rate.
The success/failure counts are 124/36 for \m{} and 82/78 for
Standard FM.

Treating rollout outcomes as independent observations,
we apply a two-sided two-proportion $z$-test with a pooled
variance estimate and no continuity correction.
Under the null hypothesis of equal aggregate success rates,
the pooled proportion is
$\hat{p}=(124+82)/(160+160)=0.64375$.
The test yields $z=4.903$ and $p=9.45\times10^{-7}$,
supporting the aggregate improvement of \m{} over Standard FM.
This analysis is conditional on the deployed policy checkpoints
and the eight evaluated tasks.

\begin{table}[t]
\centering
\caption{Main hyperparameters for real-world experiments. Hyperparameter settings not listed here
are identical to those in Tab.~\ref{tab:simulation_hyperparameters}.}
\label{tab:real-hparams}
\begin{tabular}{ll}
\toprule
Hyperparameter & Value \\
\midrule
Robot platform & Galaxea R1 Lite \\
Actuated degrees of freedom & Right arm (6) + gripper (1) \\
Control and recording rate & $15$\,Hz \\
Cameras & Head-mounted and right wrist-mounted RGB \\
Image resolution & $256 \times 256$ (pad to square, then resize) \\
Proprioception / action dimension & $7$ / $7$ \\
Language encoder & Frozen CLIP ViT-B/32 ($512$-d) \\
\midrule
TOPS warm-up / continuation epochs & $500$ / $500$ \\
Total training epochs, all methods & $1000$ \\
Optimizer updates per epoch & $250$ \\
Mixture components $M_\tau$ & $3$ \\
\midrule
Rollouts per task & $20$ \\
\bottomrule
\end{tabular}
\end{table}

\subsection{Single-Task Evaluation on Robomimic}
\label{app:robomimic_single_task}

\paragraph{Experimental setup.}
We further evaluate the scheduling component of TOPS on four Robomimic~\citep{robomimic}
tasks: \textit{Lift}, \textit{Can}, \textit{Square}, and \textit{Tool Hang}.
We train a separate policy for each task using 200 proficient-human
demonstrations.
Standard FM and our scheduling variant share the same policy backbone,
a fixed \textit{Gaussian} source $\mathcal{N}(0, 0.3^2 I)$, and a training budget
of 300 epochs from scratch.
Standard FM uses linear interpolation and explicit midpoint integration,
while our variant uses slow-then-fast scheduling with its corresponding
schedule-aware integrator.
With the \textit{Gaussian} source fixed, this comparison evaluates
slow-then-fast scheduling together with its corresponding
integrator.

\begin{table}[t]
    \centering
    \caption{
        Single-task success rates (\%) on Robomimic.
        TOPS scheduling retains the same \textit{Gaussian} source as Standard FM.
        Results are mean $\pm$ sample standard deviation across three
        evaluation seeds, each evaluated on the same 50 initial states,
        for one training seed.
        Both methods use two network evaluations per integration step.
    }
    \label{tab:robomimic_single_task}
    \small
    \setlength{\tabcolsep}{6pt}
    \begin{tabular}{llccc}
        \toprule
        Task & Method & 1 step & 5 steps & 10 steps \\
        \midrule
        Lift
        & Standard FM
        & $100.0 \pm 0.0$
        & $100.0 \pm 0.0$
        & $100.0 \pm 0.0$ \\
        & TOPS scheduling
        & $100.0 \pm 0.0$
        & $100.0 \pm 0.0$
        & $100.0 \pm 0.0$ \\
        \midrule
        Can
        & Standard FM
        & $90.7 \pm 5.0$
        & $92.7 \pm 3.1$
        & $94.0 \pm 0.0$ \\
        & TOPS scheduling
        & $\mathbf{95.3 \pm 2.3}$
        & $\mathbf{95.3 \pm 4.2}$
        & $\mathbf{96.0 \pm 2.0}$ \\
        \midrule
        Square
        & Standard FM
        & $78.7 \pm 3.1$
        & $80.7 \pm 6.1$
        & $82.0 \pm 6.9$ \\
        & TOPS scheduling
        & $\mathbf{81.3 \pm 3.1}$
        & $\mathbf{86.7 \pm 5.8}$
        & $\mathbf{88.0 \pm 4.0}$ \\
        \midrule
        Tool Hang
        & Standard FM
        & $31.3 \pm 4.2$
        & $34.0 \pm 2.0$
        & $38.7 \pm 4.2$ \\
        & TOPS scheduling
        & $\mathbf{43.3 \pm 4.2}$
        & $\mathbf{46.7 \pm 5.0}$
        & $\mathbf{47.3 \pm 2.3}$ \\
        \bottomrule
    \end{tabular}
\end{table}

\paragraph{Results and interpretation.}
As shown in Tab.~\ref{tab:robomimic_single_task}, the largest gains
occur on \textit{Tool Hang}, where one-step success increases from $31.3\%$
to $43.3\%$. The improvement persists with five and ten steps,
with margins of $12.7$ and $8.6$ percentage points, respectively.
On \textit{Square}, our variant raises one-step success from $78.7\%$ to
$81.3\%$ and improves success by $6.0$ percentage points at both
five and ten steps.
\textit{Can} also achieves higher mean success than Standard FM at every
evaluated budget.
\textit{Lift} reaches $100\%$ success with both methods at all budgets,
leaving no room for further improvement in this evaluation.

These results show higher mean success on all three non-saturated
tasks across all tested integration budgets.
Notably, with only one step, our variant already achieves higher
mean success than ten-step Standard FM on \textit{Can} ($95.3\%$ versus
$94.0\%$) and \textit{Tool Hang} ($43.3\%$ versus $38.7\%$), supporting
its effectiveness under a limited inference budget.
Together, these findings complement the multi-task experiments
by showing that slow-then-fast scheduling, combined with its
corresponding integrator, can improve single-task manipulation
while retaining the same \textit{Gaussian} source.

\subsection{Toy Experiments: Interpolation Timing for Few-Step Generation}
\label{app:toy_experiments}

\paragraph{Purpose.}
We use a controlled two-dimensional generation task to examine the
slow-then-fast scheduling choice in TOPS. The task is to transform a compact
source distribution into two separated target modes. This simple setting
makes it easy to inspect whether generated samples reach the target regions
when only one or two integration steps are available. We compare three
schedules while keeping the source and target distributions, network
architecture, and training budget fixed.

\paragraph{Task and distributions.}
We follow the main-text notation with dimension $D=2$ and flow state
$\mathbf{x}_t\in\mathbb{R}^2$, where $t\in[0,1]$ denotes generative time.
The horizontal and vertical axes in Fig.~\ref{fig:toy_exp} are the first and second
coordinates of this state, respectively. Since this is a single
unconditional task, we omit the task index and observation--language
conditioning. Source samples $\mathbf{x}_0\in\mathbb{R}^2$ and target
samples $\mathbf{x}_1\in\mathbb{R}^2$ are drawn independently from the
following distributions on $\mathbb{R}^2$:
\begin{equation}
\begin{aligned}
    p_0 &= \mathcal{N}\!\left(\mathbf{0},\,0.3^2\mathbf{I}_2\right),\\
    p_{\mathrm{data}}
    &= \tfrac{1}{2}\mathcal{N}\!\left(
        \boldsymbol{\mu}_{+},\,0.15^2\mathbf{I}_2\right)
     + \tfrac{1}{2}\mathcal{N}\!\left(
        \boldsymbol{\mu}_{-},\,0.15^2\mathbf{I}_2\right),
\end{aligned}
\label{eq:toy_distributions}
\end{equation}
where $\mathcal{N}$ denotes a \textit{Gaussian} distribution with the specified
mean and covariance, $\mathbf{0}\in\mathbb{R}^2$ is the zero vector,
$\mathbf{I}_2\in\mathbb{R}^{2\times2}$ is the identity matrix, and
$\boldsymbol{\mu}_{+}=(0,2)^\top\in\mathbb{R}^2$ and
$\boldsymbol{\mu}_{-}=(0,-2)^\top\in\mathbb{R}^2$ are the target mode
centers. Thus, generation must produce two clusters above and below the
source. The vertical coordinate is the expansion direction in this
axis-aligned setting. The star in Fig.~\ref{fig:toy_exp} marks the source mean;
individual trajectories start from samples drawn around this mean.

\paragraph{Scheduling comparison.}
We retain the main-text convention that $a_j(t)$ is a source weight.
Here, $j\in\{1,2\}$ indexes the coordinates,
$x_{t,j}\in\mathbb{R}$ denotes coordinate $j$ of $\mathbf{x}_t$, and
$a_j:[0,1]\rightarrow[0,1]$ is a differentiable scalar function with
$a_j(0)=1$ and $a_j(1)=0$. We construct the path as
\begin{equation}
\begin{aligned}
    x_{t,j} &= a_j(t)x_{0,j}+[1-a_j(t)]x_{1,j},
    \qquad j\in\{1,2\},\\
    a_1(t) &= 1-t,\qquad
    a_2(t)=(1-t)e^{\rho_2 t},
\end{aligned}
\label{eq:toy_schedule}
\end{equation}
where $\rho_2\in\{-0.5,0,0.5\}\subset\mathbb{R}$ is the dimensionless
scheduling strength for the vertical coordinate. We set $\rho_2=0$ for
\emph{linear}, $\rho_2=-0.5$ for \emph{fast-then-slow}, and $\rho_2=0.5$
for \emph{slow-then-fast (ours)}. The negative value is introduced only
as a control that reverses the scheduling direction. The horizontal
coordinate remains linear in all three cases.

These labels describe the rate of weight transfer from source to target
over generative time. Writing $\dot a_2(t)=\mathrm{d}a_2(t)/\mathrm{d}t$
for the scalar time derivative, the endpoint rate ratio is
\begin{equation}
    \frac{|\dot a_2(1)|}{|\dot a_2(0)|}
    =\frac{e^{\rho_2}}{|1-\rho_2|}.
\label{eq:toy_rate_ratio}
\end{equation}
This gives $1.00$, $0.40$, and $3.30$ for linear, fast-then-slow, and
slow-then-fast scheduling, respectively.

\paragraph{Training and generation.}
For each schedule, we train a separate velocity network
$\mathbf{v}_{\theta}:\mathbb{R}^2\times[0,1]\rightarrow\mathbb{R}^2$
with trainable parameters $\theta$, following the main-text notation.
Each network is trained from scratch using the flow-matching objective
and the corresponding path velocity
$\mathbf{u}_t=\partial_t\mathbf{x}_t\in\mathbb{R}^2$.
The network receives the current state and time, with no target-mode
label. All three networks are multilayer perceptrons with three hidden
layers of 256 units and SiLU activations. We use Adam for 20,000 updates
with batch size 1,024 and learning rate $10^{-3}$. The networks share the
same initialization and the same sequence of sampled source endpoints,
target endpoints, and uniformly sampled training times.

At evaluation, we reuse the same 30,000 source samples across
schedules and integration budgets. Each learned field is integrated
over $[0,1]$ on a uniform time grid using the two-stage
schedule-aware integrator with its corresponding schedule
coefficients. Each step requires two velocity-network evaluations.
For the linear schedule, the update reduces to two explicit
Euler half-steps.

\paragraph{Evaluation measures.}
\emph{Capture} is the percentage of generated endpoints whose Euclidean
distance to either $\boldsymbol{\mu}_{+}$ or $\boldsymbol{\mu}_{-}$ is
less than $0.6$. The dashed regions in Fig.~\ref{fig:toy_exp} indicate these capture
regions. \emph{Mid-band} is the percentage of endpoints whose vertical
coordinate has magnitude less than $1$, measuring residual mass between
the two target modes. Higher capture and lower mid-band mass indicate
better placement of samples in this task. The rightmost column in Fig.~\ref{fig:toy_exp} shows
the marginal density of the vertical coordinate, providing a complementary
view of mode location and spread.

\begin{table}[t]
\centering
\caption{Results on the two-dimensional bimodal generation task.
All entries are percentages computed from the same 30,000 source
samples reused across schedules.
Fig.~\ref{fig:toy_exp} displays the same results rounded to whole percentages.}
\label{tab:toy_schedule_results}
\small
\begin{tabular}{lcccc}
\toprule
& \multicolumn{2}{c}{1 integration step (NFE = 2)}
& \multicolumn{2}{c}{2 integration steps (NFE=4)} \\
\cmidrule(lr){2-3}\cmidrule(lr){4-5}
Schedule & Capture $\uparrow$ & Mid-band $\downarrow$
         & Capture $\uparrow$ & Mid-band $\downarrow$ \\
\midrule
Linear         & 24.87 & 22.39 & 84.16 & 0.89 \\
Fast-then-slow & 16.05 & 55.91 & 40.23 & 2.31 \\
Slow-then-fast (ours)
               & \textbf{34.10} & \textbf{12.93}
               & \textbf{98.57} & \textbf{0.73} \\
\bottomrule
\end{tabular}
\end{table}

\paragraph{Results and interpretation.}
With one integration step, slow-then-fast scheduling achieves the highest
capture, approximately $34\%$, compared with $25\%$ for linear and
$16\%$ for fast-then-slow scheduling. Fast-then-slow scheduling leaves
approximately $56\%$ of samples in the mid-band, compared with $22\%$
and $13\%$ for linear and slow-then-fast scheduling. All three schedules
still show substantial deviations from the target at this budget.

With two integration steps, slow-then-fast scheduling reaches approximately
$99\%$ capture, compared with $84\%$ for linear and $40\%$ for
fast-then-slow scheduling. The plotted vertical marginals visually show
better alignment with the target mode locations. At this budget,
mid-band mass is small for all three schedules. The remaining errors
therefore include samples that have left the central band but fall
outside the target regions; leaving the middle alone does not ensure
accurate mode placement.

These results support the slow-then-fast design for few-step generation
in this expansion setting. They are consistent with the main-text
motivation that competing target modes make early velocity prediction
ambiguous, and that a slower initial transition followed by later
acceleration can improve generation under a small integration budget.
The experiment provides complementary evidence with a bimodal target
to the \textit{Gaussian} endpoint analysis.

\subsection{Language-Based Task Identification}
\label{app:open_language}

We evaluate language-based task identification on 358 distinct instruction
formulations from the environment-D validation annotations, covering
34 known CALVIN tasks.
Table~\ref{tab:open_language_retrieval} compares canonical task-name
retrieval, leave-one-out prototype retrieval, and a learned task
classifier evaluated out of fold.

For prototype-based retrieval, we group instruction formulations by task
label and represent each task by the normalized mean of its normalized
instruction embeddings.
Each query is excluded from the support set used to construct the
prototypes for its evaluation.
This leave-one-out protocol evaluates held-out wording within the known
task set.

Given a normalized query embedding $\mathbf{e}(\ell)$, we
select the task whose prototype $\mathbf{p}_{\tau}^{(-\ell)}$
has the highest cosine similarity:
\begin{equation}
    \hat{\tau}
    = \psi_{\mathrm{proto}}(\ell)
    = \arg\max_{\tau\in\mathcal{T}}
      \left\langle
          \mathbf{e}(\ell),\mathbf{p}_{\tau}^{(-\ell)}
      \right\rangle .
    \label{eq:open_language_proto}
\end{equation}

In addition to prototype retrieval, we evaluate a learned task
classifier operating on MiniLM instruction embeddings, with the
34 known tasks as its output classes.
We use out-of-fold evaluation: each instruction is assigned to a
held-out fold and classified by a model trained on the remaining
folds.
Top-1 and top-3 accuracy are computed from these held-out predictions,
evaluating task identification from unseen instruction formulations
within the known task set.

Tab.~\ref{tab:open_language_retrieval} compares the language
encoders and task representations. We construct a prototype selector using the 384-dimensional
MiniLM-based sentence embeddings
\citep{minilm,sentencebert}
provided with CALVIN.
With CLIP, replacing canonical task names with leave-one-out
prototypes improves top-1 accuracy from $32.7\%$ to $51.4\%$.
MiniLM prototypes further increase top-1 accuracy to $81.3\%$
and achieve $98.3\%$ top-3 accuracy.
Under out-of-fold evaluation, the MiniLM-based learned classifier
achieves $98.6\%$ top-1 and $100.0\%$ top-3 accuracy.
Its reported top-1 accuracy exceeds that of MiniLM prototype
retrieval by $17.3$ percentage points, supporting accurate
identification of known tasks from held-out instruction wording.

\begin{table}[t]
    \centering
    \small
    \caption{
    Language-based task-selection accuracy over 358 distinct
    instruction formulations for 34 known tasks.
    Prototype evaluations exclude each query from its support set.
    The learned classifier is evaluated out of fold, with each
    instruction classified by a model trained on the remaining folds.
}

    \label{tab:open_language_retrieval}
    \begin{tabular}{llrr}
        \toprule
        Encoder & Task representation & Top-1 (\%) & Top-3 (\%) \\
        \midrule
        CLIP   & Canonical task names
        & 32.7 & 66.8 \\
        CLIP   & Leave-one-out prototypes
        & 51.4 & 96.1 \\
        MiniLM & Leave-one-out prototypes
        & 81.3 & 98.3 \\
                MiniLM & Learned classifier (out-of-fold)
        & \textbf{98.6} & \textbf{100.0} \\

        \bottomrule
    \end{tabular}
\end{table}

\subsection{Velocity-Prediction Risk and Schedule Design}
\label{app:raft_risk_proof}

\subsubsection{Proof of the Endpoint-Speed Relation}
\label{app:risk-theoretical-analysis}

We prove Eq.~\ref{eq:raft_spectral_endpoint_ratio} under the \textit{Gaussian} endpoint model.
The argument proceeds by deriving a direction-wise upper bound
on velocity-prediction risk, minimizing its time integral, and
evaluating the resulting schedule at the two endpoints.

\paragraph{Setup and directional variances.}
Fix a task and suppress its index $\tau$.
Let $D\in\mathbb{N}_{>0}$ denote the action-chunk dimension.
Consider independent \textit{Gaussian} random vectors
$\boldsymbol{x}_0,\boldsymbol{x}_1\in\mathbb{R}^D$:
\begin{equation}
\boldsymbol{x}_0\sim
\mathcal{N}(\boldsymbol{0},\mathbf{C}_s),
\qquad
\boldsymbol{x}_1\sim
\mathcal{N}(\boldsymbol{0},\mathbf{C}_d),
\qquad
\boldsymbol{x}_0\perp\boldsymbol{x}_1,
\label{eq:app_speed_endpoints}
\end{equation}
where $\boldsymbol{0}\in\mathbb{R}^D$ is the zero vector and
$\mathbf{C}_s,\mathbf{C}_d\in\mathbb{R}^{D\times D}$ are
symmetric positive-definite covariance matrices.
It suffices to consider centered endpoints: nonzero endpoint
means only add deterministic, time-dependent shifts to the
interpolated state and velocity, leaving their conditional
variances unchanged.

Let $\mathbf{T}\in\mathbb{R}^{D\times D}$ be the symmetric
positive-definite \textit{Gaussian} OT map for the squared Euclidean cost~\citep{formula_distri,Comput_OP_trans}. Its covariance-matching
identity and eigendecomposition are
\begin{equation}
\mathbf{T}\mathbf{C}_s\mathbf{T}=\mathbf{C}_d,
\qquad
\mathbf{T}
=\mathbf{R}\operatorname{diag}(s_1,\ldots,s_D)\mathbf{R}^{\top}.
\label{eq:app_speed_ot}
\end{equation}
Here,
$\mathbf{R}=[\boldsymbol{r}_1,\ldots,\boldsymbol{r}_D]
\in\mathbb{R}^{D\times D}$ is orthogonal,
$\boldsymbol{r}_j\in\mathbb{R}^D$ is its $j$th column,
and $s_j\in\mathbb{R}_{>0}$ is the corresponding eigenvalue,
for $j\in\{1,\ldots,D\}$.

To compare source and target variability along the same
directions, define
\begin{equation}
\boldsymbol{y}_i
:=\mathbf{R}^{\top}\boldsymbol{x}_i\in\mathbb{R}^D,
\quad i\in\{0,1\},
\qquad
\nu_j
:=\boldsymbol{r}_j^{\top}\mathbf{C}_s\boldsymbol{r}_j
\in\mathbb{R}_{>0}.
\label{eq:app_speed_coordinates}
\end{equation}
Writing $y_{i,j}\in\mathbb{R}$ for the $j$th coordinate of
$\boldsymbol{y}_i$, we have
$\operatorname{Var}(y_{0,j})=\nu_j$.
For the target coordinate,
Eq.~\ref{eq:app_speed_ot} gives
\begin{equation}
\begin{aligned}
\operatorname{Var}(y_{1,j})
&=\boldsymbol{r}_j^{\top}\mathbf{C}_d\boldsymbol{r}_j\\
&=\boldsymbol{r}_j^{\top}
\mathbf{T}\mathbf{C}_s\mathbf{T}\boldsymbol{r}_j\\
&=(s_j\boldsymbol{r}_j)^{\top}
\mathbf{C}_s(s_j\boldsymbol{r}_j)
=s_j^2\nu_j.
\end{aligned}
\label{eq:app_speed_target_variance}
\end{equation}
Thus, $s_j$ is the target-to-source standard-deviation ratio
along direction $\boldsymbol{r}_j$.

Let $\mathcal{A}$ denote the set of continuously differentiable,
nonincreasing scalar functions $a:[0,1]\to[0,1]$ satisfying
$a(0)=1$ and $a(1)=0$.
For generation time $t\in[0,1]$ and schedules
$a_1,\ldots,a_D\in\mathcal{A}$, define the interpolated state
$\boldsymbol{y}_t\in\mathbb{R}^D$ and its velocity
$\boldsymbol{w}_t:=\partial_t\boldsymbol{y}_t\in\mathbb{R}^D$
through their scalar coordinates:
\begin{equation}
\begin{aligned}
y_{t,j}
&=a_j(t)y_{0,j}+[1-a_j(t)]y_{1,j},\\
w_{t,j}
&=\dot a_j(t)(y_{0,j}-y_{1,j}).
\end{aligned}
\label{eq:app_speed_path}
\end{equation}
Here, $y_{t,j},w_{t,j}\in\mathbb{R}$, and dots denote
differentiation with respect to $t$.
Because $\mathbf{R}$ is orthogonal, predicting
$\boldsymbol{w}_t$ from $\boldsymbol{y}_t$ has the same minimum
squared error as the corresponding prediction in action space.

\paragraph{From prediction error to a direction-wise bound.}
We first identify the smallest prediction error attainable
under this endpoint model.
For a fixed $t$, let
$\boldsymbol{f}:\mathbb{R}^D\to\mathbb{R}^D$ be any measurable
predictor with
$\mathbb{E}\|\boldsymbol{f}(\boldsymbol{y}_t)\|_2^2<\infty$,
and define the conditional mean
\begin{equation}
\boldsymbol{m}_t
:=\mathbb{E}[\boldsymbol{w}_t\mid\boldsymbol{y}_t]
\in\mathbb{R}^D.
\label{eq:app_speed_conditional_mean}
\end{equation}
The residual $\boldsymbol{w}_t-\boldsymbol{m}_t$ has zero
conditional mean given $\boldsymbol{y}_t$.
Consequently, expanding the squared error and conditioning
the cross term on $\boldsymbol{y}_t$ yields
\begin{equation}
\begin{aligned}
\mathbb{E}\|
\boldsymbol{w}_t-\boldsymbol{f}(\boldsymbol{y}_t)
\|_2^2
&=
\mathbb{E}\|
\boldsymbol{w}_t-\boldsymbol{m}_t
\|_2^2\\
&\quad+
\mathbb{E}\|
\boldsymbol{m}_t-\boldsymbol{f}(\boldsymbol{y}_t)
\|_2^2.
\end{aligned}
\label{eq:app_speed_error_decomposition}
\end{equation}
The second term is nonnegative and vanishes when the predictor
equals the conditional mean.
The Bayes velocity-prediction risk
$\mathcal{R}:[0,1]\to\mathbb{R}_{\geq0}$ is therefore
\begin{equation}
\begin{aligned}
\mathcal{R}(t)
&:=
\inf_{\boldsymbol{f}}
\mathbb{E}\|
\boldsymbol{w}_t-\boldsymbol{f}(\boldsymbol{y}_t)
\|_2^2\\
&=
\mathbb{E}\|
\boldsymbol{w}_t-
\mathbb{E}[\boldsymbol{w}_t\mid\boldsymbol{y}_t]
\|_2^2\\
&=
\sum_{j=1}^D
\mathbb{E}\!\left[
\operatorname{Var}(w_{t,j}\mid\boldsymbol{y}_t)
\right].
\end{aligned}
\label{eq:app_speed_bayes_risk}
\end{equation}
The last equality follows by expanding the squared Euclidean
norm into its coordinate-wise squared residuals.
All expectations are over the independent endpoint samples.

A direction-wise bound will allow the subsequent schedule
optimization to separate across directions.
To obtain such a bound, we compare conditioning on the full
state $\boldsymbol{y}_t$ with conditioning on its single
coordinate $y_{t,j}$.
Since $y_{t,j}$ is a function of $\boldsymbol{y}_t$, the
conditional law of total variance gives
\begin{equation}
\begin{aligned}
\operatorname{Var}(w_{t,j}\mid y_{t,j})
&=
\mathbb{E}\!\left[
\operatorname{Var}(w_{t,j}\mid\boldsymbol{y}_t)
\,\middle|\, y_{t,j}
\right]\\
&\quad+
\operatorname{Var}\!\left(
\mathbb{E}[w_{t,j}\mid\boldsymbol{y}_t]
\,\middle|\, y_{t,j}
\right).
\end{aligned}
\label{eq:app_speed_total_variance}
\end{equation}
Taking expectations, the first term on the right becomes
$\mathbb{E}[\operatorname{Var}
(w_{t,j}\mid\boldsymbol{y}_t)]$, while the second term
remains nonnegative.
Summing over $j$ and using
Eq.~\ref{eq:app_speed_bayes_risk}, we obtain
\begin{equation}
\mathcal{R}(t)
\leq
\sum_{j=1}^D
\mathbb{E}\!\left[
\operatorname{Var}(w_{t,j}\mid y_{t,j})
\right].
\label{eq:app_speed_coordinate_bound}
\end{equation}
This step allows correlations between different directions;
it only uses the fact that the full state contains its
individual coordinates.

\paragraph{Evaluating the scalar conditional variance.}
We now compute each term in
Eq.~\ref{eq:app_speed_coordinate_bound}.
For this calculation, fix $t$ and $j$ and abbreviate
$a:=a_j(t)\in[0,1]$,
$b:=\dot a_j(t)\in\mathbb{R}$,
$s:=s_j\in\mathbb{R}_{>0}$, and
$\nu:=\nu_j\in\mathbb{R}_{>0}$.
The pair $(y_{t,j},w_{t,j})$ is jointly \textit{Gaussian} because both
variables are linear combinations of the independent \textit{Gaussian}
endpoints $y_{0,j}$ and $y_{1,j}$.
Using Eq.~\ref{eq:app_speed_path} and their variances gives
\begin{equation}
\begin{aligned}
\operatorname{Var}(y_{t,j})
&=a^2\nu+(1-a)^2s^2\nu
=\nu\bigl[a^2+s^2(1-a)^2\bigr],\\
\operatorname{Var}(w_{t,j})
&=b^2(\nu+s^2\nu)
=\nu b^2(1+s^2),\\
\operatorname{Cov}(w_{t,j},y_{t,j})
&=b\bigl[a\nu-(1-a)s^2\nu\bigr]\\
&=\nu b\bigl[a-s^2(1-a)\bigr].
\end{aligned}
\label{eq:app_speed_scalar_moments}
\end{equation}
The cross-endpoint covariance terms vanish by independence.
Moreover, $\operatorname{Var}(y_{t,j})>0$ throughout
$t\in[0,1]$, since $\nu,s>0$ and $a$ and $1-a$ cannot
both vanish.

To see how these moments determine the conditional variance,
define the scalar regression coefficient and residual
\begin{equation}
\beta_{t,j}
:=
\frac{\operatorname{Cov}(w_{t,j},y_{t,j})}
     {\operatorname{Var}(y_{t,j})}
\in\mathbb{R},
\qquad
\varepsilon_{t,j}
:=w_{t,j}-\beta_{t,j}y_{t,j}
\in\mathbb{R}.
\label{eq:app_speed_regression_residual}
\end{equation}
By construction,
\begin{equation}
\begin{aligned}
\operatorname{Cov}(\varepsilon_{t,j},y_{t,j})
&=
\operatorname{Cov}(w_{t,j},y_{t,j})
-\beta_{t,j}\operatorname{Var}(y_{t,j})\\
&=0.
\end{aligned}
\label{eq:app_speed_residual_covariance}
\end{equation}
The residual and $y_{t,j}$ are jointly \textit{Gaussian}, so this zero
covariance implies independence.
Conditioning on $y_{t,j}$ fixes the term
$\beta_{t,j}y_{t,j}$ and leaves only the residual uncertainty.
Hence
\begin{equation}
\begin{aligned}
\operatorname{Var}(w_{t,j}\mid y_{t,j})
&=\operatorname{Var}(\varepsilon_{t,j})\\
&=\operatorname{Var}(w_{t,j})
-2\beta_{t,j}\operatorname{Cov}(w_{t,j},y_{t,j})
+\beta_{t,j}^2\operatorname{Var}(y_{t,j})\\
&=\operatorname{Var}(w_{t,j})
-\frac{\operatorname{Cov}(w_{t,j},y_{t,j})^2}
      {\operatorname{Var}(y_{t,j})}.
\end{aligned}
\label{eq:app_speed_conditional_variance}
\end{equation}

Substituting Eq.~\ref{eq:app_speed_scalar_moments} into
Eq.~\ref{eq:app_speed_conditional_variance} gives
\begin{equation}
\begin{aligned}
\operatorname{Var}(w_{t,j}\mid y_{t,j})
&=
\nu b^2
\left[
(1+s^2)
-\frac{[a-s^2(1-a)]^2}{a^2+s^2(1-a)^2}
\right]\\
&=
\nu b^2
\frac{
(1+s^2)[a^2+s^2(1-a)^2]
-[a-s^2(1-a)]^2
}{
a^2+s^2(1-a)^2
}.
\end{aligned}
\label{eq:app_speed_variance_substitution}
\end{equation}
The numerator simplifies as
\begin{equation}
\begin{aligned}
&(1+s^2)[a^2+s^2(1-a)^2]
-[a-s^2(1-a)]^2\\
&\qquad=
s^2\bigl[a^2+2a(1-a)+(1-a)^2\bigr]
=s^2.
\end{aligned}
\label{eq:app_speed_numerator}
\end{equation}
Restoring the original notation therefore yields
\begin{equation}
\operatorname{Var}(w_{t,j}\mid y_{t,j})
=
\frac{\nu_j s_j^2\dot a_j(t)^2}
     {a_j(t)^2+s_j^2[1-a_j(t)]^2}.
\label{eq:app_speed_scalar_risk}
\end{equation}
This expression does not depend on the realized value of
$y_{t,j}$, so taking its expectation leaves it unchanged.
Combining Eqs.~\ref{eq:app_speed_coordinate_bound}
and~\ref{eq:app_speed_scalar_risk} gives the spectral upper
bound $\overline{\mathcal{R}}:
[0,1]\to\mathbb{R}_{\geq0}$:
\begin{equation}
\mathcal{R}(t)
\leq
\overline{\mathcal{R}}(t)
:=
\sum_{j=1}^D
\frac{\nu_j s_j^2\dot a_j(t)^2}
     {a_j(t)^2+s_j^2[1-a_j(t)]^2}.
\label{eq:app_speed_spectral_bound}
\end{equation}

\paragraph{Minimizing the integrated upper bound.}
The bound in Eq.~\ref{eq:app_speed_spectral_bound} expresses
the contribution of each direction using only its own schedule.
To make this separation explicit, define the positive scalar
function $g_j:[0,1]\to\mathbb{R}_{>0}$ and the functional
$J_j:\mathcal{A}\to\mathbb{R}_{\geq0}$ by
\begin{equation}
g_j(r)
:=
\frac{\nu_j s_j^2}{r^2+s_j^2(1-r)^2},
\quad r\in[0,1],
\qquad
J_j[a]
:=
\int_0^1 g_j(a(t))\dot a(t)^2\,dt.
\label{eq:app_speed_functional}
\end{equation}
The integrated bound then becomes
\begin{equation}
\int_0^1\overline{\mathcal{R}}(t)\,dt
=
\sum_{j=1}^D J_j[a_j].
\label{eq:app_speed_separation}
\end{equation}
Each term depends only on $a_j$, and each schedule is subject
to the same independent admissibility constraints.
We can therefore minimize the integrated bound by minimizing
$J_j$ separately for every direction.

Define the positive constant
\begin{equation}
L_j
:=
\int_0^1\sqrt{g_j(r)}\,dr
\in\mathbb{R}_{>0}.
\label{eq:app_speed_length}
\end{equation}
For any $a\in\mathcal{A}$, monotonicity gives
$|\dot a(t)|=-\dot a(t)$.
Together with $a(0)=1$ and $a(1)=0$, this implies
\begin{equation}
\begin{aligned}
\int_0^1
\sqrt{g_j(a(t))}\,|\dot a(t)|\,dt
&=
-\int_0^1
\sqrt{g_j(a(t))}\,\dot a(t)\,dt\\
&=
-\int_1^0\sqrt{g_j(r)}\,dr
=L_j.
\end{aligned}
\label{eq:app_speed_change_of_variable}
\end{equation}
Thus, this integral is fixed by the endpoints, independently
of the timing of the schedule.
Applying the Cauchy--Schwarz inequality~\citep{Cauchy_schwarz} on the unit time
interval gives
\begin{equation}
\begin{aligned}
J_j[a]
&=
\int_0^1
\left[\sqrt{g_j(a(t))}\,|\dot a(t)|\right]^2dt\\
&\geq
\left(
\int_0^1
\sqrt{g_j(a(t))}\,|\dot a(t)|\,dt
\right)^2
=L_j^2.
\end{aligned}
\label{eq:app_speed_lower_bound}
\end{equation}
Equality is attained when
$\sqrt{g_j(a(t))}\,|\dot a(t)|$ is constant in time.

To verify that an admissible schedule can attain this equality,
define $F_j:[0,1]\to[0,1]$ by
\begin{equation}
F_j(u)
:=
\frac{1}{L_j}\int_u^1\sqrt{g_j(r)}\,dr,
\qquad u\in[0,1].
\label{eq:app_speed_cumulative}
\end{equation}
Because $g_j$ is smooth and strictly positive,
\begin{equation}
F_j(0)=1,
\qquad
F_j(1)=0,
\qquad
F_j'(u)
=-\frac{\sqrt{g_j(u)}}{L_j}<0.
\label{eq:app_speed_cumulative_properties}
\end{equation}
Hence $F_j$ is a continuously differentiable, strictly
decreasing bijection with a continuously differentiable inverse.
Define
\begin{equation}
a_j^\star(t):=F_j^{-1}(t),
\qquad
F_j(a_j^\star(t))=t.
\label{eq:app_speed_minimizer}
\end{equation}
This schedule belongs to $\mathcal{A}$:
it decreases from $a_j^\star(0)=1$ to $a_j^\star(1)=0$.
Differentiating the second identity in
Eq.~\ref{eq:app_speed_minimizer} gives
\begin{equation}
-\frac{\sqrt{g_j(a_j^\star(t))}}{L_j}
\dot a_j^\star(t)=1,
\qquad
\sqrt{g_j(a_j^\star(t))}
\,|\dot a_j^\star(t)|=L_j.
\label{eq:app_speed_optimal_condition}
\end{equation}
It follows that $J_j[a_j^\star]=L_j^2$.
Thus, the constructed schedule attains the lower bound in
Eq.~\ref{eq:app_speed_lower_bound} and minimizes the
directional contribution to the integrated spectral bound.

\paragraph{The endpoint-speed relation.}
It remains to evaluate
Eq.~\ref{eq:app_speed_optimal_condition} at the two endpoints.
From the definition of $g_j$,
\begin{equation}
g_j(1)=\nu_j s_j^2,
\qquad
g_j(0)=\nu_j.
\label{eq:app_speed_endpoint_weights}
\end{equation}
Since the source endpoint corresponds to $a_j^\star(0)=1$
and the data endpoint to $a_j^\star(1)=0$, we obtain
\begin{equation}
\begin{aligned}
|\dot a_j^\star(0)|
&=
\frac{L_j}{\sqrt{g_j(1)}}
=\frac{L_j}{s_j\sqrt{\nu_j}},\\
|\dot a_j^\star(1)|
&=
\frac{L_j}{\sqrt{g_j(0)}}
=\frac{L_j}{\sqrt{\nu_j}}.
\end{aligned}
\label{eq:app_speed_endpoint_values}
\end{equation}
Taking their ratio and restoring the task index yields
\begin{equation}
\frac{|\dot a_{\tau j}^\star(1)|}
     {|\dot a_{\tau j}^\star(0)|}
=s_{\tau j},
\qquad j=1,\ldots,D,
\label{eq:app_speed_endpoint_ratio}
\end{equation}
which proves Eq.~\ref{eq:raft_spectral_endpoint_ratio}.
For an expansion direction, $s_{\tau j}>1$, the optimal
schedule therefore transfers interpolation weight more slowly
at the source endpoint than at the data endpoint.
This endpoint asymmetry motivates the slow-then-fast
parameterization used in the main text.

\subsubsection{Empirical Evaluation of Velocity-Prediction Risk}
\label{app:risk-validation}

We evaluate the scheduling principle in Section~\ref{sec:raft_RAFT} on LIBERO-10
using independent samples from the fitted task source $q_\tau$ and
recorded expert action chunks.
We hold the source and the original OT basis fixed, reschedule only
expansion directions, and perform the analysis without observation
conditioning.

\begin{figure}[t]
    \centering
    \includegraphics[width=\linewidth]{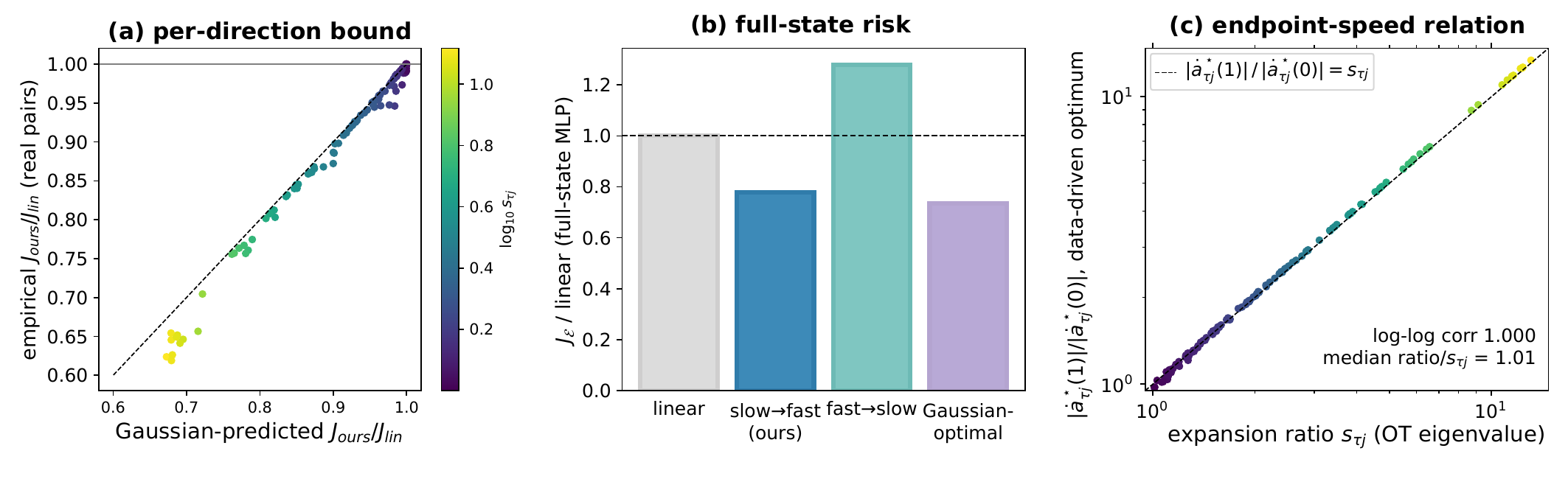}
    \vspace{-20pt}
    \caption{
    \textbf{Empirical support for the slow-then-fast scheduling principle.}
    (a) \textit{Gaussian}-predicted versus empirically estimated
    slow-then-fast/linear ratios of time-integrated directional risk terms;
    each point is an expansion direction, colored by
    $\log_{10}s_{\tau j}$.
    (b) Expansion-subspace prediction error of full-state MLP
    regressors, normalized by the linear baseline after summing over tasks.
    \textit{Gaussian}-optimal optimizes the \textit{Gaussian} bound on expansion directions.
    (c) Endpoint-speed ratios of numerical optima reconstructed
    from empirical conditional variances, with the theoretical relation
    shown as a dashed line.
    Panel (c) evaluates these reconstructed optima; the bounded TOPS
    schedule is evaluated in (a,b).
    }
    \label{fig:risk-validation}
\end{figure}

\paragraph{Agreement with the theoretical prediction.}
For each expansion direction, we estimate the conditional variance of
the velocity target given the corresponding interpolated coordinate,
and integrate it over time.
Fig.~\ref{fig:risk-validation}(a) compares the resulting
slow-then-fast/linear ratios with the \textit{Gaussian} predictions.
Across 119 expansion directions, the two sets of ratios closely agree
(correlation $0.997$). On these expansion directions, the summed integrated \textit{Gaussian}
bound decreases from $58.98$ under linear interpolation to
$48.55$ under our schedule, compared with the optimum of $45.52$,
capturing $77.5\%$ of the attainable reduction.
Summing the estimated directional risk terms over tasks and expansion
directions, our schedule reduces the total by $18.5\%$, close to the
$17.7\%$ reduction predicted by the \textit{Gaussian} analysis.
This supports the predicted scheduling benefit on real endpoint
distributions.

\paragraph{Reduced velocity-prediction error.}
We next train matched MLPs to predict velocity from the complete
interpolated state and time, using one training run per task and schedule.
Evaluation uses fresh source samples and new pairings from the same
expert-window bank.
As shown in Fig.~\ref{fig:risk-validation}(b), our schedule reduces
time-integrated prediction error on the expansion subspace by
$22.1\%$ relative to linear interpolation, whereas the fast-then-slow
control increases it by $28.0\%$.
Our schedule improves on linear interpolation in all ten tasks;
the reduction across all action coordinates is $14.2\%$.
These results support the choice of slow-then-fast timing in Section~\ref{sec:raft_RAFT}.
The measured errors include finite-model fitting error and therefore
serve as empirical proxies for the Bayes risk.

\paragraph{Consistency of the endpoint-speed relation.}
Finally, we reconstruct the optimal scalar schedules from the estimated
conditional variances using the construction in
App.~\ref{app:raft_risk_proof}.
Their endpoint-speed ratios closely follow $s_{\tau j}$
(log--log correlation $0.9997$; Fig.~\ref{fig:risk-validation}(c)),
consistent with Eq.~\ref{eq:raft_spectral_endpoint_ratio}.
This provides a numerical consistency check of the endpoint asymmetry
that motivates our bounded schedule, complementing the measured
reductions in panels (a,b).

\subsection{Derivation of the Schedule-Aware Integrator}
\label{app:integrator_details}

Given a selected source--schedule pair $(q_\tau,\mathbf{K}_\tau)$ and a fixed
policy condition $c$, we initialize $\mathbf{x}^{(0)}=\mathbf{x}_0\sim q_\tau$ and generate
an action chunk using the following two-stage schedule-aware update.
For language-based inference, $\tau=\hat\tau$ as defined
in Section~\ref{sec:raft_training_inference}.

\begin{equation}
\boxed{
\begin{aligned}
\widetilde{\mathbf{x}}^{(n+1/2)}
&=
\mathbf{x}^{(n)}
+
\bigl(\mathbf{S}_{t_n}^{\tau}
      -\mathbf{S}_{t_{n+1/2}}^{\tau}\bigr)
\bigl(\mathbf{G}_{t_n}^{\tau}\bigr)^{-1}
\mathbf{v}_{\boldsymbol{\theta}}
(\mathbf{x}^{(n)},t_n,c),
\\
\mathbf{x}^{(n+1)}
&=
\widetilde{\mathbf{x}}^{(n+1/2)}
+
\bigl(\mathbf{S}_{t_{n+1/2}}^{\tau}
      -\mathbf{S}_{t_{n+1}}^{\tau}\bigr)
\bigl(\mathbf{G}_{t_{n+1/2}}^{\tau}\bigr)^{-1}
\mathbf{v}_{\boldsymbol{\theta}}
(\widetilde{\mathbf{x}}^{(n+1/2)},t_{n+1/2},c).
\end{aligned}
}
\label{eq:ti-solver}
\end{equation}
Here $N\in\mathbb{N}_{>0}$ is the number of solver steps,
$t_n=n/N$, $h=1/N$, and $t_{n+1/2}=t_n+h/2$, for
$n=0,\ldots,N-1$.
The numerical state $\mathbf{x}^{(n)}$ and predicted midpoint
$\widetilde{\mathbf{x}}^{(n+1/2)}$ belong to $\mathbb{R}^{D}$,
where $D=Hd_a$ is the flattened action-chunk dimension defined
in the main text.
The condition $c$ remains fixed throughout integration.

\paragraph{Schedule operators.}
The matrices in Eq.~\ref{eq:ti-solver} are
\begin{equation}
\begin{aligned}
\mathbf{S}_t^\tau
&:=
(1-t)e^{-t\mathbf{K}_\tau}
\in\mathbb{R}^{D\times D},
\\
\mathbf{G}_t^\tau
&:=
-\partial_t\mathbf{S}_t^\tau
=
e^{-t\mathbf{K}_\tau}
\bigl[\mathbf{I}_D+(1-t)\mathbf{K}_\tau\bigr]
\in\mathbb{R}^{D\times D}.
\end{aligned}
\label{eq:ti-operators}
\end{equation}
Here $\mathbf{I}_D$ is the $D\times D$ identity matrix, and
$\mathbf{K}_\tau
=\mathbf{R}_\tau
\operatorname{diag}(-\rho_{\tau1},\ldots,-\rho_{\tau D})
\mathbf{R}_\tau^\top$
is the fixed scheduling matrix from Section~4.3.
The matrix $\mathbf{R}_\tau\in\mathbb{R}^{D\times D}$ is
orthogonal, and each scalar strength satisfies
$0\leq\rho_{\tau j}<\kappa<1$.
Consequently,
\begin{equation}
\bigl(\mathbf{G}_t^\tau\bigr)^{-1}
=
\mathbf{R}_\tau
\operatorname{diag}\!\left(
\left\{
\frac{e^{-t\rho_{\tau j}}}
     {1-(1-t)\rho_{\tau j}}
\right\}_{j=1}^{D}
\right)
\mathbf{R}_\tau^\top.
\label{eq:ti-inverse}
\end{equation}
Every denominator is at least $1-\kappa>0$, so the inverse
is well-defined for all $t\in[0,1]$.
This spectral expression requires no dense matrix inversion
or matrix-exponential evaluation.

\paragraph{One-step inference.}
Our reported one-step configuration sets $N=1$, with
$t_0=0$, $t_{1/2}=1/2$, and $t_1=1$.
Starting from $\mathbf{x}_0\sim q_\tau$, substituting these
times into Eq.~\ref{eq:ti-solver} gives
\begin{equation}
\boxed{
\begin{aligned}
\widetilde{\mathbf{x}}^{(1/2)}
&=
\mathbf{x}_0
+
\left[
\mathbf{I}_D-\frac{1}{2}e^{-\mathbf{K}_\tau/2}
\right]
\left(\mathbf{I}_D+\mathbf{K}_\tau\right)^{-1}
\mathbf{v}_{\boldsymbol{\theta}}(\mathbf{x}_0,0,c),
\\
\mathbf{x}^{(1)}
&=
\widetilde{\mathbf{x}}^{(1/2)}
+
\left(2\mathbf{I}_D+\mathbf{K}_\tau\right)^{-1}
\mathbf{v}_{\boldsymbol{\theta}}
\left(\widetilde{\mathbf{x}}^{(1/2)},\frac{1}{2},c\right).
\end{aligned}
}
\label{eq:ti-one-step}
\end{equation}
Here $\widetilde{\mathbf{x}}^{(1/2)}\in\mathbb{R}^{D}$
is the predicted midpoint, and
$\mathbf{x}^{(1)}\in\mathbb{R}^{D}$ is the generated action
chunk at generative time $t=1$.
The first network evaluation, at $(\mathbf{x}_0,0)$,
predicts the midpoint through the known schedule.
The second evaluation, at
$(\widetilde{\mathbf{x}}^{(1/2)},1/2)$, updates the
displacement estimate and completes the remaining half
interval.
Thus, one integration step consists of two sequential
velocity-network evaluations, with the task-specific
matrix $\mathbf{K}_\tau$ determining both update weights.

\paragraph{Derivation.}
For an endpoint pair
$\mathbf{x}_0,\mathbf{x}_1\in\mathbb{R}^{D}$, define its
displacement
$\boldsymbol{\Delta}:=\mathbf{x}_1-\mathbf{x}_0
\in\mathbb{R}^{D}$.
The path and target velocity in Eq.~(10) become
\begin{equation}
\mathbf{x}_t
=
\mathbf{x}_1-\mathbf{S}_t^\tau\boldsymbol{\Delta},
\qquad
\mathbf{u}_t
=
\mathbf{G}_t^\tau\boldsymbol{\Delta}.
\label{eq:ti-path}
\end{equation}
Thus, a velocity prediction induces the normalized
displacement estimate
\begin{equation}
\widehat{\boldsymbol{\Delta}}_{\boldsymbol{\theta}}^\tau
(\mathbf{x},t,c)
:=
\bigl(\mathbf{G}_t^\tau\bigr)^{-1}
\mathbf{v}_{\boldsymbol{\theta}}(\mathbf{x},t,c)
\in\mathbb{R}^{D}.
\label{eq:ti-displacement}
\end{equation}
For scalar times $0\leq a<b\leq1$, integrating
$\mathbf{G}_t^\tau=-\partial_t\mathbf{S}_t^\tau$ gives
\begin{equation}
\mathbf{x}_b-\mathbf{x}_a
=
\int_a^b\mathbf{G}_t^\tau\boldsymbol{\Delta}\,\mathrm{d}t
=
\bigl(\mathbf{S}_a^\tau-\mathbf{S}_b^\tau\bigr)
\boldsymbol{\Delta}.
\label{eq:ti-exact-increment}
\end{equation}
Replacing $\boldsymbol{\Delta}$ by the estimate at the
left endpoint yields a schedule-aware update over $[a,b]$.
Applying this update first over $[t_n,t_{n+1/2}]$, then
re-estimating the displacement at the predicted midpoint
and advancing over $[t_{n+1/2},t_{n+1}]$, gives
Eq.~\ref{eq:ti-solver}.

This construction integrates the known schedule exactly
while holding the displacement estimate fixed within
each half-step.
If each velocity query equals the target velocity of a
fixed endpoint pair, Eq.~\ref{eq:ti-displacement} recovers
$\boldsymbol{\Delta}$ and both updates reproduce that
pair's path exactly.
For a learned conditional field, however, squared-loss
regression targets a conditional mean displacement,
$\mathbb{E}[\boldsymbol{\Delta}\mid\mathbf{x}_t,t,c]$,
which generally varies along the trajectory.
The fixed-pair identity therefore does not imply exact
integration of the learned ODE.
When $\mathbf{K}_\tau=\mathbf{0}$, the update reduces to two
explicit Euler half-steps; for general smooth learned
fields, the composition is first-order accurate.

\subsection{Complete Training and Inference Pipeline}
\label{app:complete-pipeline}

Algorithm~\ref{alg:tops-pipeline} summarizes the complete TOPS
pipeline: warm-up and flow reversal, offline source and schedule
construction, continuation training, and action generation.
Let $\mathcal{T}$ denote the set of training tasks and
$\mathcal{D}=\{(\mathbf{x}_1,\textbf{c},\tau)\}$ the demonstration dataset.
The integers $E_{\mathrm{w}},E_{\mathrm{tr}}>0$ specify the warm-up
and continuation update budgets, respectively.
Sampling and loss expressions are shown for one example and are
averaged over minibatches in implementation.
All action-space vectors belong to $\mathbb{R}^{D}$, where
$D=Hd_a$, and all schedule matrices belong to
$\mathbb{R}^{D\times D}$.
The policy parameters
$\boldsymbol{\theta},\boldsymbol{\theta}_0,
\boldsymbol{\theta}^{\star}\in\mathbb{R}^{P}$
contain $P$ trainable entries.

Source samples are drawn independently of the demonstration
conditional on its task.
The source distributions and scheduling matrices remain fixed
after offline construction.
Inference uses the matrices
$\mathbf{S}_t^\tau$ and $\mathbf{G}_t^\tau$ defined in
Eq.~\ref{eq:ti-operators} and follows
Eq.~\ref{eq:ti-solver}.
Our default $N=1$ gives the explicit one-step update in
Eq.~\ref{eq:ti-one-step}, requiring two network evaluations.

\begin{algorithm}[!htbp]
\caption{TOPS Training and Inference Pipeline}
\label{alg:tops-pipeline}
\small
\renewcommand{\algorithmicrequire}{\textbf{Require:}}
\renewcommand{\algorithmicensure}{\textbf{Output:}}
\begin{algorithmic}[1]

\Require Demonstrations
$\mathcal D=\{(\mathbf{x}_1,c,\tau)\}$;
policy $\mathbf{v}_{\boldsymbol{\theta}}$;
update budgets $E_w,E_{\mathrm{tr}}$;
warm-up source scale $\sigma_0>0$;
mixture counts $\{M_\tau\}_{\tau\in\mathcal T}$ and rank $r$;
schedule parameters $\gamma>0$, $0<\kappa<1$;
learned source selector $\psi_{\boldsymbol{\phi}}$;
solver steps $N\geq 1$.
\Ensure Trained policy $\mathbf{v}_{\boldsymbol{\theta}^\star}$
and frozen source--schedule pairs
$\{(q_\tau,\mathbf{K}_\tau)\}_{\tau\in\mathcal T}$.

\Statex \textbf{Phase 0: Warm-up and flow reversal}
\For{$e=1,\ldots,E_w$}
    \State Sample $(\mathbf{x}_1,c,\tau)\sim\mathcal D$,
    $\mathbf{x}_0\sim\mathcal N(\mathbf{0},\sigma_0^2\mathbf{I}_D)$,
    and $t\sim\mathcal U(0,1)$.
    \State $\mathbf{x}_t^{\mathrm{lin}}
    \gets(1-t)\mathbf{x}_0+t\mathbf{x}_1$;
    $\mathbf{u}_t^{\mathrm{lin}}
    \gets\mathbf{x}_1-\mathbf{x}_0$.
    \State Update $\boldsymbol{\theta}$ using
    $\|\mathbf{v}_{\boldsymbol{\theta}}
    (\mathbf{x}_t^{\mathrm{lin}},t,c)
    -\mathbf{u}_t^{\mathrm{lin}}\|_2^2$.
\EndFor
\State $\boldsymbol{\theta}_0\gets\boldsymbol{\theta}$;
initialize $B_\tau\gets\emptyset$ for every $\tau\in\mathcal T$.
\For{each $(\mathbf{x}_1,c,\tau)\in\mathcal D$}
    \State $\widehat{\mathbf{x}}_0
    \gets\Phi^{c}_{\boldsymbol{\theta}_0,1\rightarrow 0}
    (\mathbf{x}_1)$.
    \State $B_\tau\gets B_\tau\cup\{\widehat{\mathbf{x}}_0\}$.
\EndFor

\Statex \textbf{Phase 1: Offline source and schedule construction}
\For{each $\tau\in\mathcal T$}
    \State $q_\tau
    \gets\operatorname{FitMPPCA}(B_\tau;M_\tau,r)$.
    \State Compute the full mixture covariance $\mathbf{C}_s^\tau$
    from $q_\tau$ and the expert covariance $\mathbf{C}_d^\tau$
    from task-$\tau$ action chunks.
    \State Regularize $\mathbf{C}_s^\tau$ and $\mathbf{C}_d^\tau$
    to ensure positive definiteness.
    \State $\mathbf{T}_\tau\gets
    (\mathbf{C}_s^\tau)^{-1/2}
    \bigl[
        (\mathbf{C}_s^\tau)^{1/2}
        \mathbf{C}_d^\tau
        (\mathbf{C}_s^\tau)^{1/2}
    \bigr]^{1/2}
    (\mathbf{C}_s^\tau)^{-1/2}$.
    \State Eigendecompose
    $\mathbf{T}_\tau
    =\mathbf{R}_\tau
    \operatorname{diag}(s_{\tau 1},\ldots,s_{\tau D})
    \mathbf{R}_\tau^\top$.
    \State $\rho_{\tau j}\gets
    \kappa\tanh(\gamma\log s_{\tau j})
    \mathbf{1}\{s_{\tau j}>1\}$,
    for $j=1,\ldots,D$.
    \State $\mathbf{K}_\tau\gets
    \mathbf{R}_\tau
    \operatorname{diag}(-\rho_{\tau 1},\ldots,-\rho_{\tau D})
    \mathbf{R}_\tau^\top$.
    \State Define
    $\mathbf{S}_t^\tau\gets(1-t)e^{-t\mathbf{K}_\tau}$.
    \State Define
    $\mathbf{G}_t^\tau\gets
    e^{-t\mathbf{K}_\tau}
    \bigl[\mathbf{I}_D+(1-t)\mathbf{K}_\tau\bigr]$.
\EndFor

\Statex \textbf{Phase 2: Continuation training}
\State Initialize $\boldsymbol{\theta}\gets\boldsymbol{\theta}_0$;
freeze all $q_\tau$ and $\mathbf{K}_\tau$.
\For{$e=1,\ldots,E_{\mathrm{tr}}$}
    \State Sample $(\mathbf{x}_1,c,\tau)\sim\mathcal D$;
    independently draw $\mathbf{x}_0\sim q_\tau$
    and $t\sim\mathcal U(0,1)$.
    \State $\mathbf{x}_t\gets
    \mathbf{x}_1+\mathbf{S}_t^\tau(\mathbf{x}_0-\mathbf{x}_1)$;
    $\mathbf{u}_t\gets
    \mathbf{G}_t^\tau(\mathbf{x}_1-\mathbf{x}_0)$.
    \State Update $\boldsymbol{\theta}$ using
    $\|\mathbf{v}_{\boldsymbol{\theta}}(\mathbf{x}_t,t,c)
    -\mathbf{u}_t\|_2^2$.
\EndFor
\State $\boldsymbol{\theta}^\star\gets\boldsymbol{\theta}$.

\Statex \textbf{Phase 3: Language-based source selection and inference}
\State Given instruction $\ell$ and the original policy condition $c$.
\State $\hat{\tau}\gets\psi_{\boldsymbol{\phi}}(\ell)$.
\State Select $q_{\hat{\tau}}$ and its associated
$\mathbf{K}_{\hat{\tau}}$;
sample $\mathbf{x}^{(0)}\sim q_{\hat{\tau}}$.
\State Set $t_n=n/N$ for $n=0,\ldots,N$.
\For{$n=0,\ldots,N-1$}
    \State $t_{n+1/2}\gets(t_n+t_{n+1})/2$.
    \State $\mathbf{v}^{(n)}\gets
    \mathbf{v}_{\boldsymbol{\theta}^\star}
    (\mathbf{x}^{(n)},t_n,c)$.
    \State $\widetilde{\mathbf{x}}^{(n+1/2)}
    \gets\mathbf{x}^{(n)}
    +\bigl(
        \mathbf{S}_{t_n}^{\hat{\tau}}
        -\mathbf{S}_{t_{n+1/2}}^{\hat{\tau}}
    \bigr)
    \bigl(\mathbf{G}_{t_n}^{\hat{\tau}}\bigr)^{-1}
    \mathbf{v}^{(n)}$.
    \State $\mathbf{v}^{(n+1/2)}\gets
    \mathbf{v}_{\boldsymbol{\theta}^\star}
    (\widetilde{\mathbf{x}}^{(n+1/2)},t_{n+1/2},c)$.
    \State $\mathbf{x}^{(n+1)}
    \gets\widetilde{\mathbf{x}}^{(n+1/2)}
    +\bigl(
        \mathbf{S}_{t_{n+1/2}}^{\hat{\tau}}
        -\mathbf{S}_{t_{n+1}}^{\hat{\tau}}
    \bigr)
    \bigl(\mathbf{G}_{t_{n+1/2}}^{\hat{\tau}}\bigr)^{-1}
    \mathbf{v}^{(n+1/2)}$.
\EndFor
\State \Return $\mathbf{x}^{(N)}$
as the generated normalized action chunk.

\end{algorithmic}
\end{algorithm}

For deployment, $\mathbf{x}^{(N)}$ is reshaped into an
$H\times d_a$ action array and unnormalized.
The controller executes the configured action window,
then repeats Phase~3 using the updated observation.

\end{document}